\documentclass{article}
\usepackage[final,main]{neurips_2026}

\usepackage[utf8]{inputenc}
\usepackage[T1]{fontenc}
\usepackage{hyperref}
\hypersetup{
  hidelinks,
  hypertexnames=false,
  pdftitle={Oracle Supervision Transfers for Hyperparameter Prediction in Model-Based Image Denoising},
  pdfauthor={Jianmin Liao, Lixin Shen, Yuesheng Xu},
  pdfsubject={Accepted at NeurIPS 2026}
}
\usepackage{url}
\usepackage{booktabs}
\usepackage{array}
\usepackage{multirow}
\usepackage{amsfonts}
\usepackage{amsmath}
\usepackage{amssymb}
\usepackage{amsthm}
\usepackage{mathtools}
\usepackage{bm}
\usepackage{microtype}
\usepackage{xcolor}
\usepackage{graphicx}
\usepackage{float}
\usepackage{needspace}
\usepackage{placeins}
\usepackage{tikz}
\usetikzlibrary{positioning,arrows.meta,fit,backgrounds,calc}

\newtheorem{theorem}{Theorem}

\theoremstyle{definition}
\theoremstyle{plain}

\newcommand{\E}{\mathbb E}

\title{Oracle Supervision Transfers for Hyperparameter Prediction in Model-Based Image Denoising}

\author{%
  Jianmin Liao\thanks{Corresponding author.} \\
  Department of Mathematics \\
  Syracuse University \\
  \texttt{jliao21@syr.edu}
  \And
  Lixin Shen \\
  Department of Mathematics \\
  Syracuse University \\
  \texttt{lshen03@syr.edu}
  \And
  Yuesheng Xu \\
  Department of Mathematics \\
  and Statistics \\
  Old Dominion University \\
  \texttt{y1xu@odu.edu}
}

\begin{document}

\maketitle

\begin{abstract}
Hyperparameter prediction is a critical practical bottleneck for model-based image denoisers, ranging from classical TV/TGV variational solvers to modern diffusion-based models such as DiffPIR.
While existing learned predictors can achieve near-oracle performance, this approach scales poorly: each new configuration conventionally requires its own oracle-labeled training set, and each label requires a hierarchical grid search evaluated against clean ground truth.
We therefore ask whether oracle supervision collected on source configurations can transfer to target configurations with few or no target oracle labels.
We propose HyperDn, a single configuration-conditioned predictor that pools oracle supervision across source configurations and predicts heterogeneous hyperparameters for new denoiser--noise configurations.
In a cross-paradigm experiment, HyperDn transfers from relatively cheap TV/TGV variational sources to more expensive diffusion-based DiffPIR.
With $2$ target labels for predictor fitting after shared configuration calibration, it comes within $0.90$\,dB of the oracle and matches the $64$-label Afkham baseline trained from scratch while using $1/32$ as many target labels.
In addition, transfer remains effective when oracle labels are defined by SSIM or LPIPS. We also observe clear transfer gains from variational models to NLM and CBM3D, despite substantial differences in hyperparameter semantics.
Without any target oracle labels, HyperDn also reaches near-oracle PSNR on two unseen mixtures of seen noise types, on transfer from mostly $96{\times}96$ source images to targets as large as roughly $1024{\times}768$, and under simultaneous shifts in image domain and diffusion prior, from WED natural images with an ImageNet-trained DiffPIR prior to FFHQ faces with an FFHQ-trained prior.
Together, these results show that transferable oracle supervision can replace repeated per-configuration labeling with a shared upfront cost amortized across configurations, plus a small marginal labeling cost for the evaluated targets.
\end{abstract}

\section{Introduction}

Model-based denoisers include variational methods such as TV~\citep{rudin1992tv} and TGV~\citep{bredies2010tgv}, nonlocal patch-based methods such as \mbox{NLM~\citep{buades2005nlm}} and \mbox{CBM3D~\citep{dabov2007cbm3d}}, and diffusion-based methods such as DiffPIR~\citep{zhu2023denoising}. Their reconstruction quality depends strongly on image-dependent hyperparameters. Learned predictors can approach oracle selection for one fixed inverse-problem configuration~\citep{afkham2021learning}, but their supervision is costly: every label evaluates a hierarchical hyperparameter sweep against clean ground truth, and the process is repeated for each denoising configuration. For example, one DiffPIR call on a $256{\times}256$ image takes about $10$ seconds in our setup.

The number of configurations, $N_{\mathrm{cfg}}$, is also large: each specifies a denoiser, a noise setting, and, where applicable, a data term. There are numerous existing denoisers, including variational, nonlocal, and diffusion-based ones, while noise types include Gaussian, Poisson, black-and-white impulse, RGB impulse, and their mixtures. The design space keeps expanding as new methods and noise mixtures appear.
The total cost therefore scales as
\[
\underbrace{t_{\mathrm{call}}}_{\text{sec/call}}
\;\times\; \underbrace{K_{\mathrm{sweep}}}_{\text{candidates/image}}
\;\times\; \underbrace{N_{\mathrm{img}}}_{\text{training images}}
\;\times\; \underbrace{N_{\mathrm{cfg}}}_{\text{configurations}}.
\]

We ask whether oracle labels collected on source configurations can reduce the label need of a new target configuration. Transfer is plausible for two reasons. First, image properties such as noise strength, texture, and edges affect hyperparameter selection across denoisers. Second, hyperparameters can have similar semantics while taking different numerical values. For example, TV and TGV explicitly fit $\hat{\bm{x}}=\arg\min_{\bm{u}}D(\bm{u},\bm{y})+\lambda R(\bm{u})$, while DiffPIR analogously uses $\lambda$ to balance data consistency against its diffusion prior. Supervision from relatively cheap TV/TGV configurations can therefore inform hyperparameter prediction for the more expensive DiffPIR. We introduce HyperDn, a single predictor trained on labels pooled across source configurations. A shared encoder extracts image properties, while a configuration-conditioned slot readout handles different parameter sets and denoiser-specific ranges.

We evaluate transfer across denoisers, evaluation metrics, noise types, image domains, and image resolutions. Transferring from TV/TGV to DiffPIR, HyperDn reaches $30.23$\,dB with two target labels, matching the $64$-label Afkham per-configuration CNN predictor with $1/32$ as many labels. Transfer also holds under SSIM and LPIPS: one target label gives better mean scores than Afkham with $16$ labels under both objectives. For NLM and CBM3D, HyperDn achieves the highest mean PSNR at every tested label budget despite their different hyperparameter semantics; with two target labels, HyperDn outperforms Afkham by $3.73$ and $3.95$\,dB, respectively.
Without any target oracle labels, HyperDn transfers zero-shot from WED natural images to FFHQ faces while DiffPIR simultaneously switches from an ImageNet-trained to an FFHQ-trained prior, achieving near-oracle PSNR with a gap of $0.273$\,dB. With no target labels, HyperDn also stays within ${\sim}0.2$\,dB of the oracle on two unseen mixtures of seen noise types and transfers from mostly $96{\times}96$ source images to targets up to roughly $1024{\times}768$. On $512{\times}768$ Kodak24, oracle labeling costs roughly $40{\times}$ more denoiser time than at $96{\times}96$, making it especially valuable to reuse oracle labels from low-resolution images.

\begin{figure}[!htbp]
\centering
\includegraphics[width=\textwidth]{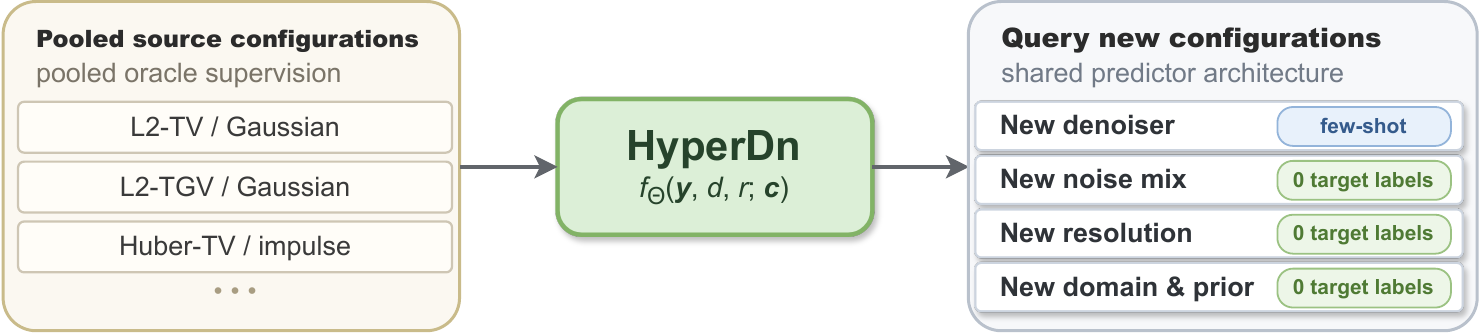}
\caption{\textbf{HyperDn: one pool of oracle labels for many configurations.} Oracle labels from multiple source configurations are pooled to train a shared conditional predictor $f_{\Theta}(\bm{y},\,d,\,r;\,\bm{c})$. HyperDn reuses this supervision for few-shot transfer to new denoisers (DiffPIR, NLM, and CBM3D) and zero-shot transfer to unseen mixtures of seen noise types, larger resolutions, and simultaneous shifts in image domain and diffusion prior. Here, $\bm{y}$ is the degraded image, $d$ is the data-term tag, $r$ is the prior identifier, and $\bm{c}$ encodes noise attributes and observation metadata only when provided.}
\label{fig:teaser}
\end{figure}

\Needspace{5\baselineskip}
\paragraph{Contributions.}
\begin{itemize}
  \item We formulate oracle hyperparameter labels for image denoising as reusable supervision pooled across configurations.
  \item We introduce HyperDn, a single configuration-conditioned predictor for denoisers with different hyperparameters.
  \item We show few-shot transfer from variational models to DiffPIR, NLM, and CBM3D, including denoisers with substantially different hyperparameter semantics. Two DiffPIR labels for predictor fitting match the $64$-label Afkham baseline, and we observe similar transfer gains with SSIM- or LPIPS-defined labels. Without any target oracle labels, HyperDn also transfers zero-shot to unseen mixtures of seen noise types, to larger resolutions, and from WED natural images to FFHQ faces while DiffPIR simultaneously switches from an ImageNet-trained to an FFHQ-trained diffusion prior.
\end{itemize}

\section{HyperDn}
\label{sec:method}

\noindent
Given a degraded image and a denoising configuration, HyperDn (Figure~\ref{fig:architecture}) predicts the hyperparameters needed to reconstruct that image. We represent configuration-specific hyperparameters in a padded vector. A single configuration-conditioned predictor then shares information across configurations while preserving each denoiser's numerical range. It is trained on oracle labels pooled across the source configurations.

\subsection{Problem setting: oracle-supervised hyperparameter prediction}

\paragraph{Configuration and hyperparameter vector.}
We describe each configuration by a prior $R$, represented by a regularizer or denoiser, and, where applicable, a data term $D$. Here, $D$ ranges over L2 for Gaussian noise, Huber for Gaussian or impulse noise and their mixtures, and negative-log-likelihood-style (NLL) data terms for Poisson or Poisson--Gaussian observations; $R$ ranges over TV, TGV, DiffPIR, NLM, and CBM3D; observation noise spans Gaussian, Poisson, impulse, and mixtures thereof. A denoising configuration is defined as a tuple $(d, r; \bm{c})$, where $d$ is the data-term identifier, $r$ is the prior identifier, and $\bm{c}$ encodes noise attributes and observation metadata, such as camera calibration parameters, if available; $\bm{c}$ is empty otherwise.

For each configuration, we pad its hyperparameter vector to $\bm{p} \in \mathbb{R}^{L_{\max}}$, where $L_{\max}$ is the largest supported vector length. The pair $(d,r)$ determines the ordered active entries and the remaining padding. The Huber data term contributes the threshold $\delta$; L2 and NLL contribute none. TV and DiffPIR use the prior weight $\lambda$; TGV uses $\lambda$ together with the second-order to first-order weight ratio $\gamma$. Thus, for example, Huber-TV uses $(\lambda,\delta)$ and Huber-TGV uses $(\lambda,\gamma,\delta)$, followed by padding when needed.

\paragraph{Training and test data.}
The dataset consists of tuples $(\bm{y}, \bm{x}, d, r, \bm{c})$, where $\bm{x}$ is the clean image; each tuple is paired with an oracle hyperparameter target $\bm{p}^\star$. Let $\mathcal{R}_{d,r}(\bm{y};\bm{p})$ denote the reconstruction for $\bm{y}$ under configuration $(d,r)$ with hyperparameters $\bm{p}$ and required observation metadata implicit, $\mathcal{P}_{d,r}(\bm y,\bm x)$ the finite, nonempty set of valid hyperparameter vectors evaluated by the oracle search for this image pair, and $Q$ a quality criterion to be maximized. $Q$ can be any scalar quality measure computed for a reconstruction--reference image pair, including PSNR, SSIM~\citep{wang2004ssim}, and negative LPIPS~\citep{zhang2018lpips}; each choice defines its own oracle labels. The oracle target maximizes $Q$ against the clean image:
\begin{equation}
    \bm{p}^\star \in
    \operatorname*{arg\,max}_{\bm{p}\in\mathcal{P}_{d,r}(\bm y,\bm x)}
    Q\!\left(\mathcal{R}_{d,r}(\bm{y};\bm{p}),\,\bm{x}\right).
\end{equation}

\paragraph{Predictor input/output.}
HyperDn is a learned predictor $f_\Theta$ with parameters $\Theta$. Given configuration $(d, r; \bm{c})$ and degraded image $\bm{y}$, $f_\Theta$ produces a padded hyperparameter vector
\begin{equation}
    \hat{\bm{p}} = f_{\Theta}(\bm{y},\,d,\,r;\,\bm{c}) \in \mathbb{R}^{L_{\max}},
\end{equation}
in solver-interface units, matching $\bm p^\star$ after any cache conversion. The head predicts encoded slots $\hat{\bm q}$ and trains against $\widetilde{\bm q}=\operatorname{enc}_{d,r}(\bm p^\star)$. The configuration-specific decoder gives $\hat{\bm p}=\kappa_{d,r}(\hat{\bm q})$, applying $10^{q_j}$ to log-scale slots and leaving linear slots unchanged. The reconstruction is $\mathcal R_{d,r}(\bm y;\hat{\bm p})$.

\subsection{Architecture: configuration-conditioned hyperparameter prediction}
\label{sec:shared_slot}

\begin{figure*}[!t]
\centering
\includegraphics[width=\textwidth]{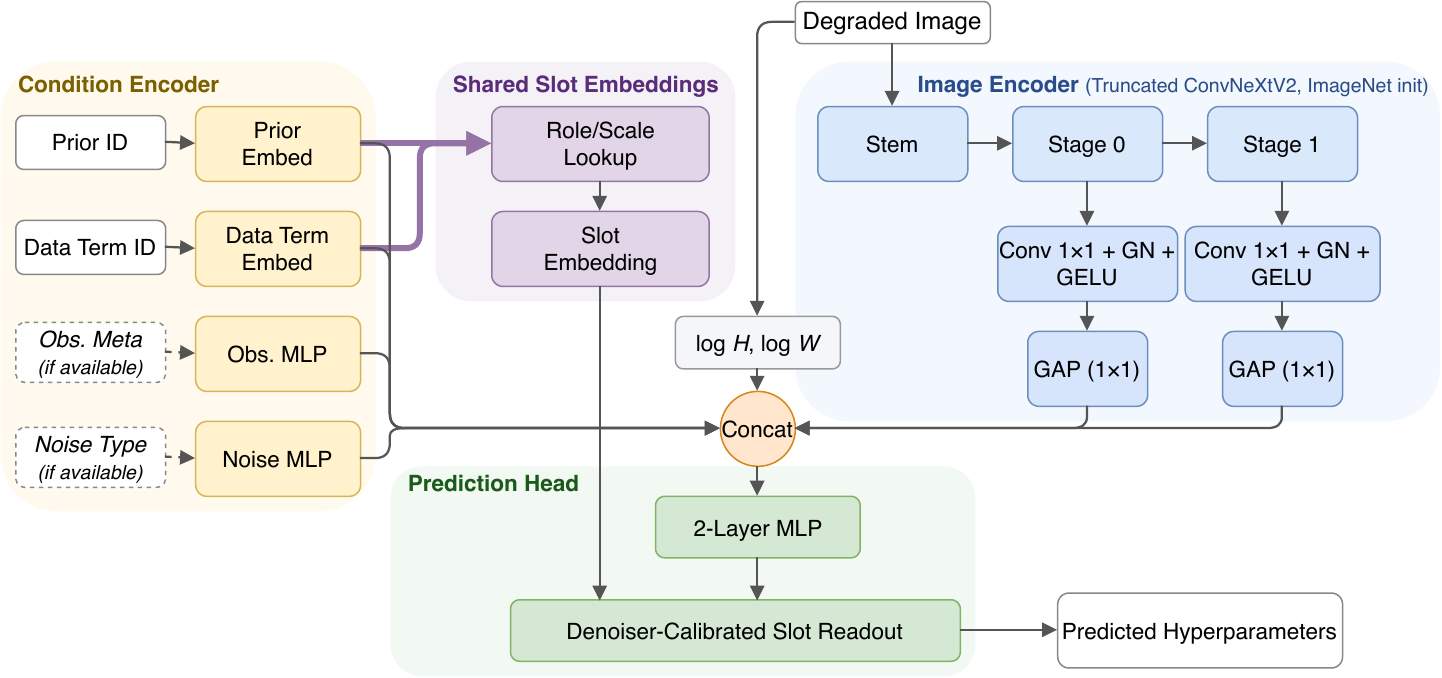}
\caption{HyperDn architecture. A shared image encoder built on a truncated ConvNeXtV2 (stem and stages 0--1; FCMAE-pretrained, then fine-tuned on ImageNet-22K and ImageNet-1K) extracts features from the degraded image. A configuration-conditioned output head, comprising a condition encoder, shared slot embeddings, and a denoiser-calibrated slot readout, produces the padded hyperparameter vector for the queried configuration; padding slots are ignored in the loss.}
\label{fig:architecture}
\end{figure*}

HyperDn has two parts: a shared encoder and an output head. The encoder uses one set of weights for all configurations, producing a feature $\bm{h}$ from the image and configuration $(d, r; \bm{c})$. The output head is designed to share hyperparameter roles across configurations while still distinguishing how each denoiser uses those roles. For example, $\lambda$ plays a prior-weight role---the weight balancing the prior against data fidelity---in TV, TGV, and DiffPIR, allowing supervision of this role to be shared across denoisers. Each denoiser uses its prior weight $\lambda$ in its own numerical range. Configurations also activate different sets of hyperparameters (e.g., L2-TV activates only $\lambda$, while Huber-TGV activates $\delta$, $\lambda$, and $\gamma$), and the head predicts each configuration's active parameters. HyperDn therefore represents the output as padded \emph{slots}: each coordinate of the encoded vector $\bm q$ is called a slot, and under $(d, r)$ a slot is either active (it holds a hyperparameter) or padding. The output head is a denoiser-calibrated slot readout that maps $\bm{h}$ and per-slot metadata to $\hat{\bm q}$. Figure~\ref{fig:architecture} shows the architecture of HyperDn.

\paragraph{Image and configuration encoder.}
The image encoder uses the ConvNeXtV2-base \texttt{fcmae\_ft\_in22k\_in1k} checkpoint~\citep{deng2009imagenet,woo2023convnextv2}, truncated to its stem and stages~0--1. Our ablation shows lower PSNR with either fewer or more stages (Appendix~\ref{sec:exp_ablation}). For each retained stage, a $1{\times}1$ convolution projects the spatial features to $128$ channels, adaptive average pooling forms a $12{\times}12$ grid, and averaging that grid produces a per-stage vector. These per-stage vectors are concatenated with $\log H$ and $\log W$ to form the image embedding, where $H$ and $W$ are the height and width of the input degraded image. The condition encoder produces four embeddings: data-term and prior embeddings of $d$ and $r$, respectively, and two optional MLP encodings of $\bm{c}$---one for noise attributes if available (a multi-hot categorical code flagging which noise types are present, e.g., Gaussian alone or black-and-white impulse $+$ Gaussian) and one for observation metadata if available. We provide \emph{observation metadata} only for Poisson--Gaussian source configurations, using per-image calibrated acquisition parameters such as photon count and read-noise standard deviation. All other configurations receive no observation metadata. If observation metadata or the noise type is unavailable, the corresponding MLP receives a fixed default vector. The four condition embeddings are concatenated with the image embedding, which includes $\log H$ and $\log W$. A 2-layer MLP then maps this concatenation to a feature $\bm{h}$ used by the denoiser-calibrated slot readout described below.

\paragraph{Denoiser-calibrated slot readout.}
Our readout is designed to share supervision across configurations with different hyperparameter sets, roles, and numerical ranges. All slots use a shared scoring function that combines $\bm h$ with their role and scale embeddings. Slots with the same role, such as $\lambda$ in TV and TGV, also share a role embedding even when their vector positions differ. To accommodate different numerical ranges for the same role across denoisers, we additionally transform the shared scores using denoiser-specific scales and offsets. For configuration $(d,r)$, the role/scale lookup (Figure~\ref{fig:architecture}) assigns each slot $j$ a role tag $\rho_j\in\{\delta,\lambda,\gamma,\texttt{UNKNOWN},\text{pad}\}$ and a scale tag $\tau_j$ ($\log_{10}$ for $\delta$, $\lambda$, and the two unseen principal scalars; linear for $\gamma$; none for padding). The \texttt{UNKNOWN} role is unused in source pretraining and activated for the unseen NLM/CBM3D targets. Learned embedding matrices $\mathbf{E}_{\mathrm{role}}$ and $\mathbf{E}_{\mathrm{scale}}$ give the slot embedding
\begin{equation*}
    \bm{e}_j \;=\; \mathbf{E}_{\mathrm{role}}[\rho_j] + \mathbf{E}_{\mathrm{scale}}[\tau_j].
\end{equation*}
The readout first computes an intermediate scalar score $s_j$ for predicting the hyperparameter in slot $j$. The matrix $\mathbf U$ and vectors $\bm w,\bm v$ are learned coefficients shared across all slots. The projection $\mathbf U\bm h+\bm w$ combines a linear map and a bias, and its inner product with $\bm e_j$ makes the score depend on the slot's role and scale. The term $\bm v^\top\bm h$ adds a shared linear contribution from $\bm h$. We then scale and shift $s_j$ using a learned denoiser-specific scale $g_{d,r,j}$ and offset $b_{d,r,j}$, and add a linear shortcut with slot-specific weight vector $\bm a_j$:
\begin{equation}
    \label{eq:slot-readout}
    s_j=(\mathbf U\bm h+\bm w)^\top\bm e_j+\bm v^\top\bm h,
    \qquad
    \hat{q}_j=
    \underbrace{g_{d,r,j}s_j + b_{d,r,j}}_{\text{denoiser-calibrated shared score}}
    + \underbrace{\bm{a}_j^\top\bm{h}}_{\text{linear shortcut}}.
\end{equation}
For log-scale slots, the decoder returns the raw parameter $\hat p_j=10^{\hat q_j}$; for linear slots, $\hat p_j=\hat q_j$. Our ablation study in Appendix~\ref{sec:exp_ablation_head} demonstrates the effectiveness of the proposed denoiser-calibrated slot readout.

\subsection{Pooled supervision}
\label{sec:pooled_supervision}

HyperDn is trained on oracle-labeled samples pooled across all source configurations. For each sample, the network predicts the full padded encoded vector $\hat{\bm q}\in\mathbb{R}^{L_{\max}}$, but the loss penalizes only the slots active for that sample's $(d,r)$---for example, $\lambda$ for TV; $\delta$ and $\lambda$ for Huber-TV; $\lambda$ and $\gamma$ for TGV.
The regression loss and training details are provided in Appendix~\ref{app:impl_details}.

\subsection{Theoretical Analysis of Transfer Error}
\label{sec:transfer_theory_main}

We analyze the error between predicted encoded hyperparameters and oracle labels on the target task.
We derive a three-term bound: source error and prediction-error discrepancies across data and tasks.
Let $f$ be the trained predictor mapping observations and supplied task information to encoded hyperparameters, with its weights and inference settings fixed. For source tasks $k=1,\ldots,K$ ($1\le K<\infty$) and target task $T$, let $f_k(x),f_T(x)$ be its outputs and $q_k^\star(x),q_T^\star(x)$ the oracle labels. Let $P_k,P_T$ be the source and target data distributions, with image samples $X_S^{(k)}\sim P_k$ and $X_T\sim P_T$, respectively. The pooled source and target root mean squared prediction errors (RMSEs) are
\begin{equation}
 \begin{aligned}
 \varepsilon_S(f)&=\left(\sum_k\alpha_k\E_{P_k}\|f_k(X_S^{(k)})-q_k^\star(X_S^{(k)})\|^2\right)^{1/2},\\
 \varepsilon_T(f)&=\left(\E_{P_T}\|f_T(X_T)-q_T^\star(X_T)\|^2\right)^{1/2},
 \end{aligned}
 \label{eq:predictor_transfer_risk}
\end{equation}
with fixed source weights $\alpha_k>0$ satisfying $\sum_k\alpha_k=1$.
Let $\Pi(P_k,P_T)$ be the set of joint distributions of the source and target samples with marginals $P_k$ and $P_T$, respectively.

\begin{subequations}\label{eq:relative_risk_shift}
In our notation, $\Delta_{\rm data}(f)$ measures changes in prediction error as the predictor generalizes to new image
distributions under the source tasks:
\begin{equation}
 \Delta_{\rm data}(f)=\sqrt{\sum_k\alpha_k\inf_{\pi\in\Pi(P_k,P_T)}\E_{\pi}\bigl\|
 [f_k(X_T)-f_k(X_S^{(k)})]-[q_k^\star(X_T)-q_k^\star(X_S^{(k)})]\bigr\|^2}.
\end{equation}
$\Delta_{\rm task}(f)$ measures changes in prediction error as the same predictor generalizes to the
target task on the same target image:
\begin{equation}
 \Delta_{\rm task}(f)=\left(\sum_k\alpha_k\E_{P_T}\bigl\|
 [f_T(X_T)-f_k(X_T)]-[q_T^\star(X_T)-q_k^\star(X_T)]\bigr\|^2\right)^{1/2}.
\end{equation}
\end{subequations}

\begin{theorem}[Transfer error of a fixed predictor]\label{thm:transfer_decomposition}
Under the definitions above and the conditions in Appendix~\ref{sec:transfer_definitions},
\begin{equation}
 \varepsilon_T(f)\le\varepsilon_S(f)+\Delta_{\rm data}(f)+\Delta_{\rm task}(f).
 \label{eq:transfer_decomposition_main}
\end{equation}
\end{theorem}

The proof is given in Appendix~\ref{app:transfer_theory}.

\section{Experiments}
\label{sec:experiments}

\subsection{Experimental setups}
\label{sec:setup}

\textbf{Goal.} We evaluate whether HyperDn can handle configurations not seen during source training.
We study few-shot transfer to DiffPIR, NLM, and CBM3D; transfer under SSIM and LPIPS objectives; and zero-shot transfer across noise mixtures and resolutions, and under simultaneous shifts in image domain and diffusion prior. \textbf{Training Setup.} For each image--configuration pair, the oracle label is the hyperparameter vector found by hierarchical grid search to maximize the chosen quality criterion. The PSNR-based DiffPIR, NLM/CBM3D, and cross-resolution studies start from the non-DiffPIR source checkpoint; FFHQ evaluation follows additional adaptation on WED DiffPIR labels. Mixed-noise transfer uses the two corresponding single-noise sources, while each SSIM/LPIPS study uses two WED TV/TGV sources labeled under its objective. \textbf{Evaluation Setup.} We report PSNR and its gap to the oracle, and SSIM or LPIPS for the metric-specific studies. Unless otherwise stated, learned-method results are reported as mean$\pm$std over $5$ random seeds.
Full experimental details are provided in the appendices.

\subsection{Few-shot transfer to DiffPIR}
\label{sec:exp_cross_paradigm}

We evaluate few-shot transfer from cheaper TV/TGV source labels to DiffPIR~\citep{zhu2023denoising} on $256{\times}256$ crops from the Waterloo Exploration Database (WED)~\citep{ma2017waterloo}.

We compare three predictors across $\{1,2,4,16,32,64,128\}$ DiffPIR target labels: \textbf{HyperDn} initialized from the source checkpoint, \textbf{ImageNet init}, and an \textbf{Afkham}-style per-configuration CNN~\citep{afkham2021learning}. ImageNet init shares HyperDn's architecture, ImageNet-pretrained backbone initialization, target-label subsets, and finetuning protocol, but omits source training. All methods share the same DiffPIR sampling schedule and oracle construction. Budgets count predictor-fitting labels after shared configuration calibration (Appendix~\ref{app:diffpir_transfer}).

\begin{figure}[!htb]
\centering
\includegraphics[width=\linewidth]{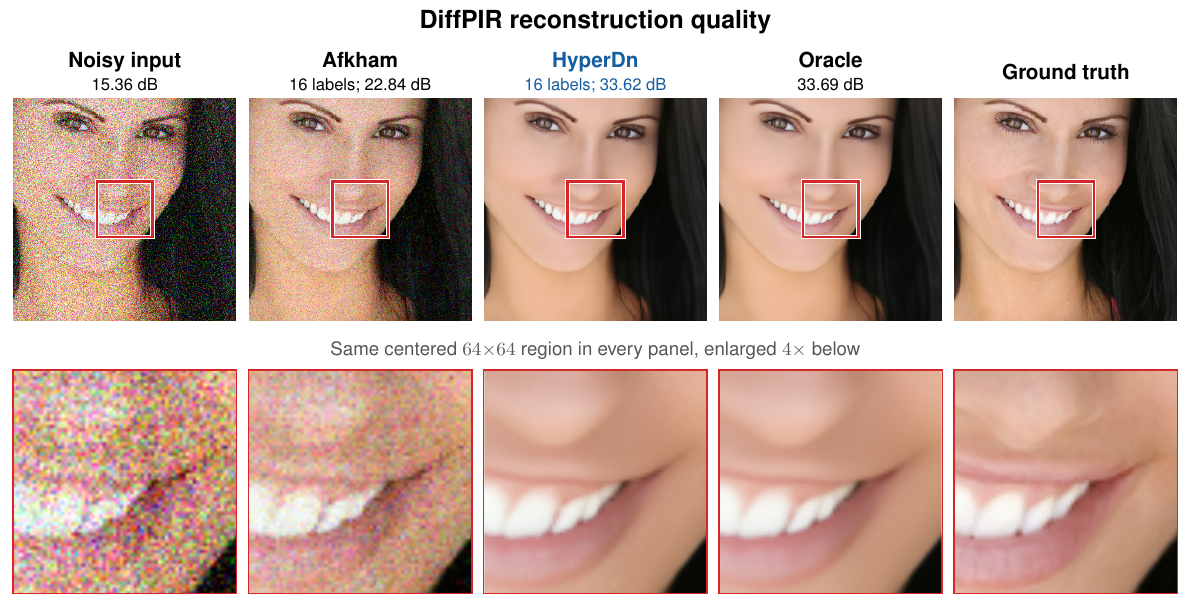}
\caption{DiffPIR reconstructions with $16$ target labels. Bottom: centered $64{\times}64$ crops at $4{\times}$ magnification.}
\label{fig:diffpir_qual}
\end{figure}

HyperDn has the highest mean at every label budget (Figure~\ref{fig:cross_paradigm_diffpir}; full results in Appendix~\ref{app:diffpir_transfer}). With $1$ label, its gap to the $31.13$\,dB oracle is $1.76$\,dB, versus $8.07$\,dB for Afkham and $6.15$\,dB for ImageNet init. With $2$ labels, HyperDn reaches $30.23$\,dB ($0.90$\,dB below the oracle), matching Afkham at $64$ labels ($30.28$\,dB) with $1/32$ of its label budget. Its gap narrows to $0.51$ and $0.10$\,dB with $4$ and $32$ labels, respectively. Source training adds $4.39$ and $2.50$\,dB over ImageNet init with $1$ and $2$ labels, respectively.
In Figure~\ref{fig:diffpir_qual}, both predictors use $16$ target labels. Afkham leaves substantial residual noise and trails the oracle by $10.85$\,dB. HyperDn is just $0.07$\,dB below the oracle and nearly visually indistinguishable from it even in the $4{\times}$ enlarged crop.

This benefit also extends to $256{\times}256$ crops from CBSD68 and Urban100: with one target label, HyperDn exceeds Afkham trained with $16$ labels by $1.68$ and $0.99$\,dB in mean PSNR, respectively (Appendix~\ref{app:additional_generalization}, Table~\ref{tab:diffpir_cross_dataset_full}).

\begin{figure}[!htb]
\centering
\includegraphics[width=0.88\linewidth]{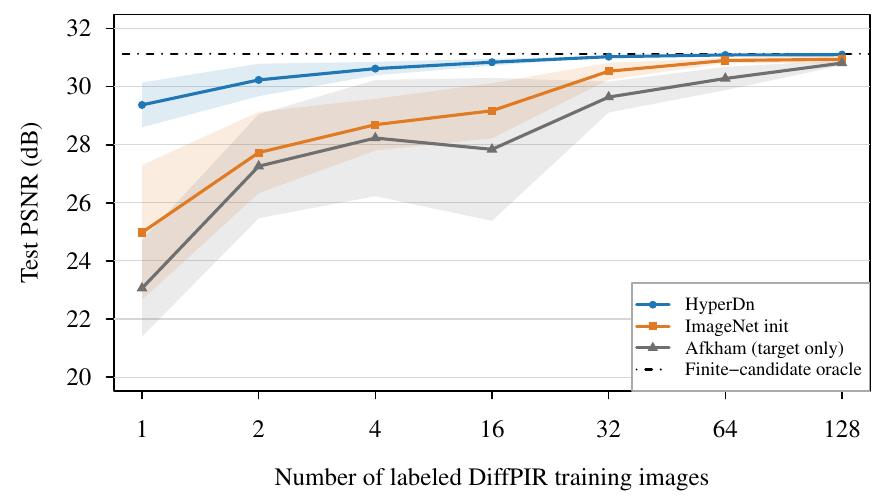}
\caption{Few-shot DiffPIR transfer on WED $256{\times}256$ crops ($100$ iterations). Curves show mean test PSNR against target-label budget; shading shows $\pm$std. Oracle: $31.13$\,dB (dash-dot).}
\label{fig:cross_paradigm_diffpir}
\end{figure}

\paragraph{Transfer under structural and perceptual objectives.}
To demonstrate that HyperDn is agnostic to the evaluation criterion, we evaluate transfer from WED TV/TGV sources to DiffPIR under SSIM and LPIPS objectives. HyperDn has the best mean at every tested budget under both objectives (Table~\ref{tab:perceptual_transfer}); with one target label, it achieves better mean scores than Afkham with $16$ labels for both SSIM ($\mathbf{.821}$ vs.\ $.789$) and LPIPS ($\mathbf{.227}$ vs.\ $.246$).

\begin{table}[!htbp]
\centering
\setlength{\tabcolsep}{2pt}
\caption{Transfer from WED TV/TGV sources to DiffPIR under SSIM and LPIPS objectives, with $N$ target training labels. Oracles: $.8911$ SSIM and $.1490$ LPIPS.}
\label{tab:perceptual_transfer}
\begin{tabular*}{\linewidth}{@{\extracolsep{\fill}}lrrrrrrr@{}}
\toprule
Method & $N{=}1$ & $2$ & $4$ & $16$ & $32$ & $64$ & $128$ \\
\midrule
\multicolumn{8}{c}{\textbf{SSIM $\uparrow$}} \\
HyperDn & \textbf{.821\textpm.055} & \textbf{.838\textpm.065} & \textbf{.877\textpm.007} & \textbf{.881\textpm.006} & \textbf{.885\textpm.002} & \textbf{.886\textpm.002} & \textbf{.888\textpm.002} \\
ImageNet init & .552\textpm.047 & .639\textpm.093 & .674\textpm.045 & .791\textpm.050 & .866\textpm.017 & .881\textpm.004 & .883\textpm.009 \\
Afkham & .537\textpm.115 & .649\textpm.095 & .698\textpm.079 & .789\textpm.018 & .838\textpm.024 & .863\textpm.013 & .872\textpm.003 \\
\midrule
\multicolumn{8}{c}{\textbf{LPIPS $\downarrow$}} \\
HyperDn & \textbf{.227\textpm.024} & \textbf{.206\textpm.005} & \textbf{.204\textpm.008} & \textbf{.177\textpm.006} & \textbf{.170\textpm.006} & \textbf{.166\textpm.005} & \textbf{.164\textpm.005} \\
ImageNet init & .387\textpm.021 & .339\textpm.041 & .320\textpm.021 & .246\textpm.030 & .193\textpm.013 & .177\textpm.006 & .167\textpm.007 \\
Afkham & .407\textpm.052 & .332\textpm.045 & .301\textpm.040 & .246\textpm.017 & .219\textpm.022 & .199\textpm.012 & .182\textpm.003 \\
\bottomrule
\end{tabular*}
\end{table}

\subsection{Zero-shot transfer to mixed noise}
\label{sec:exp_composition}

We test whether HyperDn handles unseen noise mixtures when trained only on single-noise labels.
The number of possible noise mixtures grows combinatorially with the number of noise types, so labeling every mixture is expensive.

For each unseen mixed-noise target, all three methods use the same source data---the two single-noise source training sets for the target's noise types---and evaluate directly on the mixture. The two targets are black-and-white (bw) impulse $+$ Gaussian with Huber-TV and R/G/B (rgb) impulse $+$ Gaussian with Huber-TGV. \textbf{HyperDn} pools both training sets into one. \textbf{Afkham-mean} averages predicted hyperparameters of two Afkham predictors trained independently on the two source sets. \textbf{Mean-opt} computes one mean oracle hyperparameter vector per single-noise training set, then averages the two.
These baselines test whether averaging single-noise predictions or optima is sufficient.

\begin{figure}[!htbp]
\centering
\setlength{\tabcolsep}{1.5pt}
\newcommand{\qualimg}[1]{\includegraphics[width=0.155\linewidth]{#1}}
\begin{tabular}{@{}cccccc@{}}
{\scriptsize Degraded} & {\scriptsize Clean} & {\scriptsize Oracle} & {\scriptsize \textbf{HyperDn}} & {\scriptsize Afkham-mean} & {\scriptsize Mean-opt} \\
\qualimg{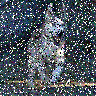} &
\qualimg{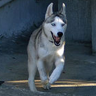} &
\qualimg{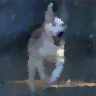} &
\qualimg{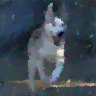} &
\qualimg{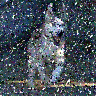} &
\qualimg{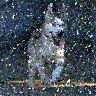} \\
{\scriptsize 11.51\,dB} & {\scriptsize --} & {\scriptsize 24.67\,dB} & {\scriptsize \textbf{24.57\,dB}} & {\scriptsize 15.33\,dB} & {\scriptsize 16.68\,dB} \\
\qualimg{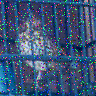} &
\qualimg{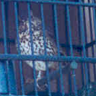} &
\qualimg{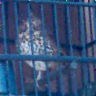} &
\qualimg{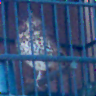} &
\qualimg{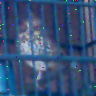} &
\qualimg{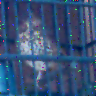} \\
{\scriptsize 17.38\,dB} & {\scriptsize --} & {\scriptsize 29.82\,dB} & {\scriptsize \textbf{29.76\,dB}} & {\scriptsize 25.25\,dB} & {\scriptsize 24.75\,dB} \\
\end{tabular}
\caption{Reconstruction samples of zero-shot transfer to mixed noise. Top row: bw impulse $+$ Gaussian with Huber-TV. Bottom row: rgb impulse $+$ Gaussian with Huber-TGV. Both averaging baselines (Afkham-mean, Mean-opt) leave residual noise; HyperDn matches the oracle on both targets.}
\label{fig:composition}
\end{figure}

HyperDn stays within $0.21$ / $0.09$\,dB of the oracle on the two targets (Table~\ref{tab:exp_a_composition}, Figure~\ref{fig:composition}), while Afkham-mean and Mean-opt trail by $5$--$7$\,dB. This shows that HyperDn's zero-shot mixed-noise transfer is not a trivial consequence of averaging single-noise predictors or optima.

\begin{table}[!htbp]
\centering
\caption{Zero-shot transfer to mixed-noise targets with known categorical composition codes but no observation metadata. PSNR (dB) $\uparrow$; Mean-opt and Oracle are deterministic.}
\label{tab:exp_a_composition}
\begin{tabular}{lcccc}
\toprule
Noise type (Denoiser) & Afkham-mean & Mean-opt & HyperDn & Oracle \\
\midrule
bw impulse $+$ Gaussian (Huber-TV)   & $17.44 \pm 0.92$ & $18.95$ & $\mathbf{24.31 \pm 0.07}$ & $24.52$ \\
rgb impulse $+$ Gaussian (Huber-TGV) & $19.68 \pm 0.86$ & $17.91$ & $\mathbf{25.00 \pm 0.06}$ & $25.09$ \\
\bottomrule
\end{tabular}
\end{table}

\subsection{Zero-shot cross-resolution transfer}
\label{sec:exp_cross_dataset}

\begin{figure}[!htb]
\centering
\begin{minipage}[t]{0.28\linewidth}
  \centering
  {\small\textbf{Training sets} (STL10 $96{\times}96$, WED $256{\times}256$ crops)}\\[8pt]
  \begin{tikzpicture}[baseline=(current bounding box.center)]
    \node[inner sep=0pt, rotate=-10]
      at (0.55,-0.36) {\includegraphics[width=2.0cm]{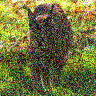}};
    \node[inner sep=0pt, rotate=5]
      at (0.13,-0.06) {\includegraphics[width=2.0cm]{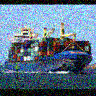}};
    \node[inner sep=0pt, rotate=-4]
      at (-0.31,0.30) {\includegraphics[width=2.0cm]{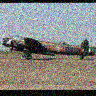}};
  \end{tikzpicture}
\end{minipage}%
\begin{minipage}[t][3.6cm][c]{0.17\linewidth}\centering
  \begin{tikzpicture}[baseline=(current bounding box.center)]
    \path[use as bounding box] (-1.05, -0.30) rectangle (1.05, 0.65);
    \node[font=\scriptsize\sffamily, text=black!60] at (0, 0.48) {HyperDn};
    \node[font=\scriptsize\sffamily, text=black!60] at (0, 0.18) {no target labels};
    \draw[-{Stealth[length=2.4mm, width=1.9mm, round]},
          line width=1.3pt, black!55, line cap=round]
          (-0.95, -0.15) -- (0.95, -0.15);
  \end{tikzpicture}
\end{minipage}%
\begin{minipage}[t]{0.55\linewidth}
  \centering
  {\small\mbox{\textbf{Test set} (Kodak24, $512{\times}768$)}}\\[4pt]
  \begin{minipage}[c]{0.36\linewidth}\centering
    \includegraphics[width=\linewidth]{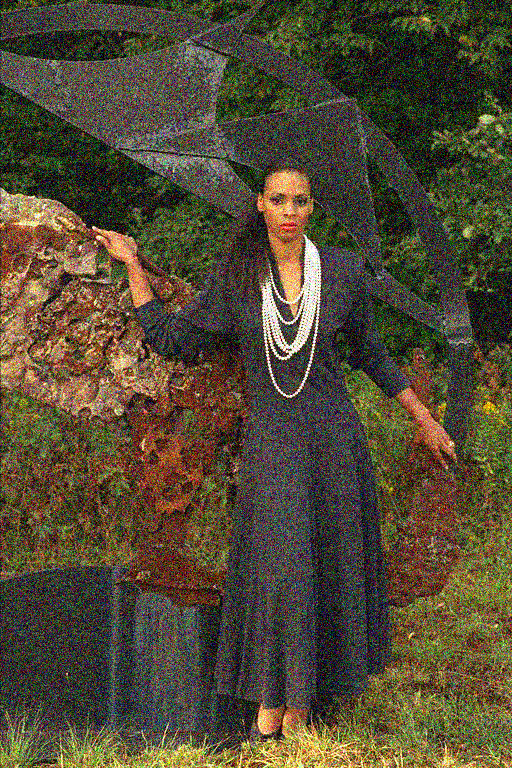}\\[2pt]
    {\scriptsize noisy input $\bm{y}$\\ Gap to oracle: $8.08$\,dB}
  \end{minipage}\hfill
  \begin{minipage}[c]{0.28\linewidth}\centering
    \begin{tikzpicture}[baseline=(current bounding box.center)]
      \path[use as bounding box] (-0.92, -0.40) rectangle (0.92, 0.55);
      \node[font=\scriptsize\sffamily, text=black!60]
        at (0, 0.40) {HyperDn$(\bm{y})\!\to\!\hat{\bm{p}}$};
      \node[font=\scriptsize\sffamily, text=black!60]
        at (0, 0.10) {TGV$(\bm{y},\hat{\bm{p}})\!\to\!\hat{\bm{x}}$};
      \draw[-{Stealth[length=2.4mm, width=1.9mm, round]},
            line width=1.3pt, black!55, line cap=round]
            (-0.80, -0.22) -- (0.80, -0.22);
    \end{tikzpicture}
  \end{minipage}\hfill
  \begin{minipage}[c]{0.36\linewidth}\centering
    \includegraphics[width=\linewidth]{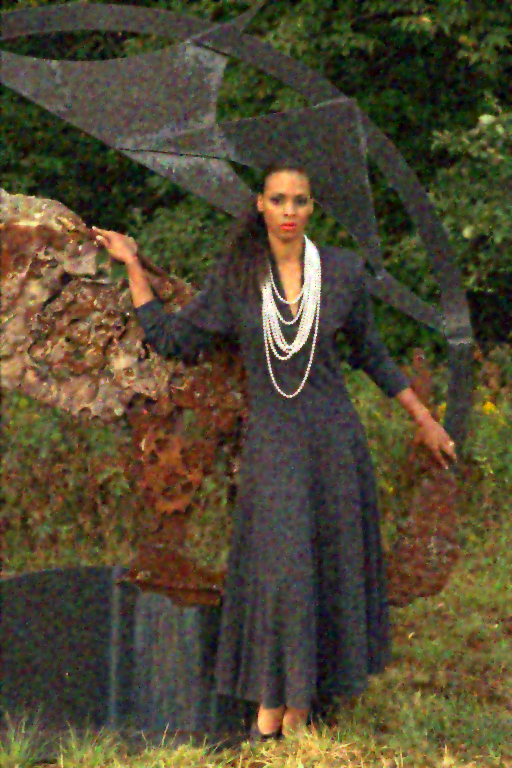}\\[2pt]
    {\scriptsize output $\hat{\bm{x}}$\\ Gap to oracle: $0.02$\,dB}
  \end{minipage}
\end{minipage}
\caption{Zero-shot cross-resolution transfer. HyperDn is trained on small-image source training sets and applied directly to larger test images without any target labels. Given the noisy input $\bm{y}$, HyperDn predicts $\hat{\bm{p}}$, which the L2-TGV denoiser then uses to produce the reconstruction image $\hat{\bm{x}}$.}
\label{fig:cross_dataset}
\end{figure}

We test whether HyperDn, source-pretrained jointly on the small-image source training sets of \S\ref{sec:setup} (STL10~\citep{coates2011stl10} $96{\times}96$ images and WED $256{\times}256$ crops, spanning multiple denoiser/noise configurations), stays near-oracle on larger test images without any target labels. A $512{\times}768$ Kodak24 oracle label costs roughly $40{\times}$ more denoiser time than a $96{\times}96$ STL10 label, so transferring from cheap small-image labels to expensive large-image targets is practically valuable. Figure~\ref{fig:cross_dataset} illustrates this cross-resolution transfer.
We evaluate three configurations on Kodak24~\citep{kodak1999}; for L2-TV with Gaussian we additionally evaluate on CBSD68~\citep{martin2001bsds} and Urban100~\citep{huang2015urban100}. HyperDn stays near-oracle in every evaluated cell (Table~\ref{tab:cross_dataset}), showing zero-shot transfer from small source images to larger target images.

\begin{table}[!htb]
\centering
\setlength{\tabcolsep}{4pt}
\caption{Zero-shot cross-resolution transfer. HyperDn is the jointly pretrained checkpoint of \S\ref{sec:setup} (STL10 $96{\times}96$ images and WED $256{\times}256$ crops), applied to each target dataset without finetuning or observation metadata. ``Gap to oracle'' is the mean over test images of (oracle PSNR $-$ HyperDn PSNR).}
\label{tab:cross_dataset}
\begin{tabular}{llccc}
\toprule
Configuration & Noise & Dataset & Resolution & Gap to oracle $\downarrow$ \\
\midrule
Huber-TV & bw impulse & Kodak24 & $512{\times}768$ & $0.06\!\pm\!0.06$ \\
\midrule
L2-TGV & Gaussian & Kodak24 & $512{\times}768$ & $0.02\!\pm\!0.01$ \\
\midrule
\multirow{3}{*}{L2-TV}     & \multirow{3}{*}{Gaussian}
                  & CBSD68          & $321{\times}481$       & $0.01\!\pm\!0.0029$ \\
 &                & Kodak24         & $512{\times}768$       & $0.27\!\pm\!0.0030$ \\
 &                & Urban100        & $\sim\!1024{\times}768$ & $0.02\!\pm\!0.0045$ \\
\bottomrule
\end{tabular}
\end{table}

\subsection{Unseen nonlocal denoisers and parameter roles}
\label{sec:additional_generalization}
These experiments show that HyperDn transfers effectively even when source and target hyperparameters have substantially different semantics. We evaluate few-shot transfer from variational source configurations to NLM~\citep{buades2005nlm} and CBM3D~\citep{dabov2007cbm3d}, neither seen during source training. The source hyperparameters control the balance or shape of explicit data-fidelity and regularization terms. In contrast, NLM's principal hyperparameter $h$ controls the decay of nonlocal patch-similarity weights and thus each patch's contribution to averaging. CBM3D's principal hyperparameter $\sigma_{\rm psd}$ specifies the assumed noise standard deviation that governs its collaborative filtering.

We adapt on $N$ labels from a $128$-image WED $256{\times}256$ pool and evaluate on $50$ test images outside this pool. ImageNet init uses the same architecture without source supervision; Afkham is a per-configuration CNN trained from scratch. Table~\ref{tab:nonlocal_transfer} reports all seven budgets, and Appendix~\ref{app:additional_generalization} gives the oracle-search details.

\begin{table}[!htbp]
\centering
\setlength{\tabcolsep}{2pt}
\caption{Few-shot transfer from variational sources to previously unseen nonlocal denoisers. PSNR (dB) $\uparrow$; oracle PSNR is $27.64$\,dB for NLM and $30.98$\,dB for CBM3D.}
\label{tab:nonlocal_transfer}
\begin{tabular*}{\linewidth}{@{\extracolsep{\fill}}lrrrrrrr@{}}
\toprule
Method & $N{=}1$ & $2$ & $4$ & $16$ & $32$ & $64$ & $128$ \\
\midrule
\multicolumn{8}{c}{\textbf{NLM}} \\
HyperDn & \textbf{24.32\textpm1.04} & \textbf{25.64\textpm1.03} & \textbf{26.41\textpm.48} & \textbf{27.18\textpm.14} & \textbf{27.31\textpm.14} & \textbf{27.47\textpm.07} & \textbf{27.53\textpm.02} \\
ImageNet init & 20.94\textpm.68 & 22.89\textpm1.98 & 21.75\textpm.62 & 24.23\textpm1.42 & 26.89\textpm.37 & 27.32\textpm.09 & 27.42\textpm.09 \\
Afkham & 21.04\textpm.81 & 21.91\textpm.93 & 23.00\textpm1.14 & 25.57\textpm.55 & 26.48\textpm.22 & 27.07\textpm.15 & 27.30\textpm.03 \\
\midrule
\multicolumn{8}{c}{\textbf{CBM3D}} \\
HyperDn & \textbf{27.09\textpm2.65} & \textbf{28.49\textpm1.29} & \textbf{29.25\textpm1.61} & \textbf{30.58\textpm.16} & \textbf{30.81\textpm.06} & \textbf{30.90\textpm.02} & \textbf{30.93\textpm.00} \\
ImageNet init & 23.43\textpm2.51 & 25.94\textpm2.23 & 26.25\textpm2.29 & 29.70\textpm.53 & 30.65\textpm.09 & 30.88\textpm.01 & 30.90\textpm.04 \\
Afkham & 22.42\textpm2.03 & 24.54\textpm2.41 & 26.16\textpm1.95 & 28.83\textpm.70 & 29.95\textpm.29 & 30.49\textpm.11 & 30.64\textpm.08 \\
\bottomrule
\end{tabular*}
\end{table}

Despite this substantial semantic shift, HyperDn outperforms both controls at matched budgets in the low-shot regime and achieves the highest mean PSNR at all seven budgets for both denoisers.

\subsection{Zero-shot image-domain and diffusion-prior shift}
\label{sec:ffhq_prior_shift}
After source pretraining and adaptation on $128$ WED DiffPIR labels generated with the ImageNet prior, we directly evaluate on $50$ FFHQ images~\citep{karras2019stylegan} using an FFHQ-trained diffusion prior unseen by the predictor. No FFHQ image or FFHQ-prior oracle label trains or adapts HyperDn. The fixed-$\lambda$ control is computed from the same $128$ WED training labels by geometric averaging, with no tuning on FFHQ. HyperDn reaches $32.553$\,dB, $0.273$\,dB below the oracle and $1.345$\,dB above the fixed control (Table~\ref{tab:ffhq_prior}).

\begin{table}[!htbp]
\centering
\setlength{\tabcolsep}{4pt}
\caption{Zero-shot DiffPIR transfer from WED natural images to FFHQ faces using an FFHQ-trained prior in place of the ImageNet-trained prior.}
\label{tab:ffhq_prior}
\begin{tabular}{@{}lrrr@{}}
\toprule
Metric & Oracle & Fixed $\lambda$ (WED) & HyperDn (zero-shot) \\
\midrule
PSNR (dB) $\uparrow$ & $32.826$ & $31.208\!\pm\!.169$ & $\mathbf{32.553\!\pm\!.010}$ \\
\bottomrule
\end{tabular}
\end{table}

We also test transfer from synthetic to real camera noise on SIDD sRGB using Huber--TV. With two target labels, HyperDn reaches $33.73$\,dB mean PSNR, compared with $31.80$\,dB for ImageNet init using the same labels (Appendix~\ref{app:additional_generalization}).

\section{Related work}
\label{sec:related_work}

\paragraph{Hyperparameter selection.}
Classical rules include the L-curve~\citep{hansen1992lcurve} and generalized cross-validation~\citep{golub1979gcv}. Imaging methods predict fixed-model parameters~\citep{afkham2021learning}, learn spatial TGV weights~\citep{vu2025tgv}, tune plug-and-play methods~\citep{wei2020tfpnp}, or use bilevel learning~\citep{delosreyes2017bilevel}. HyperDn transfers oracle supervision across models. Our Gaussian L2-TV comparison gives Monte-Carlo SURE~\citep{ramani2008sure} the true noise standard deviation (Appendix~\ref{sec:exp_classical}).

\paragraph{Learned hyperparameter optimization.}
FSBO~\citep{wistuba2021fsbo}, OptFormer~\citep{chen2022optformer}, and ranking-weighted Gaussian-process ensembles~\citep{feurer2018rgpe} transfer past searches to propose target trials and use target-task responses during search. HyperDn predicts $\hat{\bm p}$ directly from the image and configuration, with no target oracle evaluations in zero-shot transfer.

\paragraph{Meta-learning.}
MAML~\citep{finn2017maml} and Reptile~\citep{nichol2018reptile} learn initializations for rapid task adaptation. HyperDn instead trains a single configuration-conditioned predictor on pooled oracle labels across source configurations, without task-specific adaptation during pretraining.

\paragraph{Conditional parameter generation.}
Hypernetworks~\citep{ha2016hypernetworks} and dynamic-filter networks~\citep{debrabandere2016dynamic} generate network weights or input-conditioned filters; HyperRecon~\citep{wang2022hyperrecon} maps hyperparameters to reconstruction-network weights. HyperDn instead uses its shared predictor to select denoiser hyperparameters in one forward pass, retaining the existing reconstruction procedure.

\section{Conclusion and discussion}
\label{sec:discussion}

We demonstrate that oracle hyperparameter supervision transfers across denoising configurations. HyperDn, a single configuration-conditioned predictor trained on oracle labels pooled across source configurations, reduces expensive target labeling through few-shot and zero-shot transfer. The gains persist across different parameter semantics and evaluation criteria. Our analysis bounds the prediction RMSE of a fixed predictor by its source RMSE and mismatches between prediction and oracle-label changes across data and tasks. A shared source oracle bank can thus amortize its construction cost across compatible targets, reducing the marginal labeling cost of extending model-based denoising.

\paragraph{Limitations.}
Transfer to new denoisers and parameter roles requires few-shot adaptation. Mixed-noise tests use mixtures of seen types, and cross-resolution evaluation only covers the reported datasets and image sizes. Source-bank construction incurs an upfront cost.

\paragraph{Broader impacts.}
Reusing oracle supervision can reduce the repeated search cost of evaluating new regularizers or data terms. This may make model-based denoising research more accessible to groups with limited computing resources. Wider deployment of denoising may also aid privacy-sensitive surveillance or forensic enhancement.

\clearpage
\begin{ack}
This work was supported in part by National Science Foundation grants DMS-2208385 and DMS-2208386. The authors thank Syracuse University for providing computational resources through OrangeGrid.

The authors declare no competing interests.
\end{ack}

\begingroup
\small
\bibliographystyle{plainnat}
\bibliography{references,references_camera_ready}

\begin{thebibliography}{41}
\providecommand{\natexlab}[1]{#1}
\providecommand{\url}[1]{\texttt{#1}}
\expandafter\ifx\csname urlstyle\endcsname\relax
  \providecommand{\doi}[1]{doi: #1}\else
  \providecommand{\doi}{doi: \begingroup \urlstyle{rm}\Url}\fi

\bibitem[Abdelhamed et~al.(2018)Abdelhamed, Lin, and Brown]{abdelhamed2018sidd}
Abdelrahman Abdelhamed, Stephen Lin, and Michael~S. Brown.
\newblock A high-quality denoising dataset for smartphone cameras.
\newblock In \emph{Proceedings of the IEEE Conference on Computer Vision and
  Pattern Recognition (CVPR)}, pages 1692--1700, 2018.
\newblock \doi{10.1109/CVPR.2018.00182}.
\newblock URL
  \url{https://openaccess.thecvf.com/content_cvpr_2018/html/Abdelhamed_A_High-Quality_Denoising_CVPR_2018_paper.html}.

\bibitem[Boyd et~al.(2011)Boyd, Parikh, Chu, Peleato, and
  Eckstein]{boyd2011admm}
Stephen Boyd, Neal Parikh, Eric Chu, Borja Peleato, and Jonathan Eckstein.
\newblock Distributed optimization and statistical learning via the alternating
  direction method of multipliers.
\newblock \emph{Foundations and Trends in Machine Learning}, 3\penalty0
  (1):\penalty0 1--122, 2011.
\newblock \doi{10.1561/2200000016}.

\bibitem[Bredies et~al.(2010)Bredies, Kunisch, and Pock]{bredies2010tgv}
Kristian Bredies, Karl Kunisch, and Thomas Pock.
\newblock Total generalized variation.
\newblock \emph{SIAM Journal on Imaging Sciences}, 3\penalty0 (3):\penalty0
  492--526, 2010.
\newblock \doi{10.1137/090769521}.

\bibitem[Buades et~al.(2005)Buades, Coll, and Morel]{buades2005nlm}
Antoni Buades, Bartomeu Coll, and Jean-Michel Morel.
\newblock A non-local algorithm for image denoising.
\newblock In \emph{2005 IEEE Computer Society Conference on Computer Vision and
  Pattern Recognition (CVPR)}, volume~2, pages 60--65, 2005.
\newblock \doi{10.1109/CVPR.2005.38}.

\bibitem[Chen et~al.(2022)Chen, Song, Lee, Wang, Zhang, Dohan, Kawakami,
  Kochanski, Doucet, Ranzato, Perel, and de~Freitas]{chen2022optformer}
Yutian Chen, Xingyou Song, Chansoo Lee, Zi~Wang, Qiuyi Zhang, David Dohan,
  Kazuya Kawakami, Greg Kochanski, Arnaud Doucet, Marc'Aurelio Ranzato, Sagi
  Perel, and Nando de~Freitas.
\newblock Towards learning universal hyperparameter optimizers with
  transformers.
\newblock In \emph{Advances in Neural Information Processing Systems
  (NeurIPS)}, volume~35, pages 32053--32068, 2022.
\newblock URL
  \url{https://proceedings.neurips.cc/paper_files/paper/2022/hash/cf6501108fced72ee5c47e2151c4e153-Abstract-Conference.html}.

\bibitem[Coates et~al.(2011)Coates, Lee, and Ng]{coates2011stl10}
Adam Coates, Honglak Lee, and Andrew~Y. Ng.
\newblock An analysis of single-layer networks in unsupervised feature
  learning.
\newblock In \emph{Proceedings of the Fourteenth International Conference on
  Artificial Intelligence and Statistics}, volume~15 of \emph{Proceedings of
  Machine Learning Research}, pages 215--223. PMLR, 2011.
\newblock URL \url{https://proceedings.mlr.press/v15/coates11a.html}.

\bibitem[Dabov et~al.(2007)Dabov, Foi, Katkovnik, and
  Egiazarian]{dabov2007cbm3d}
Kostadin Dabov, Alessandro Foi, Vladimir Katkovnik, and Karen Egiazarian.
\newblock Color image denoising via sparse {3D} collaborative filtering with
  grouping constraint in luminance--chrominance space.
\newblock In \emph{2007 IEEE International Conference on Image Processing
  (ICIP)}, pages I--313--I--316, 2007.
\newblock \doi{10.1109/ICIP.2007.4378954}.

\bibitem[De~Brabandere et~al.(2016)De~Brabandere, Jia, Tuytelaars, and
  Van~Gool]{debrabandere2016dynamic}
Bert De~Brabandere, Xu~Jia, Tinne Tuytelaars, and Luc Van~Gool.
\newblock Dynamic filter networks.
\newblock In \emph{Advances in Neural Information Processing Systems
  (NeurIPS)}, volume~29, pages 667--675, 2016.
\newblock URL
  \url{https://papers.nips.cc/paper/2016/hash/8bf1211fd4b7b94528899de0a43b9fb3-Abstract.html}.

\bibitem[{De los Reyes} et~al.(2017){De los Reyes}, Sch{\"o}nlieb, and
  Valkonen]{delosreyes2017bilevel}
J.~C. {De los Reyes}, Carola-Bibiane Sch{\"o}nlieb, and Tuomo Valkonen.
\newblock Bilevel parameter learning for higher-order total variation
  regularisation models.
\newblock \emph{Journal of Mathematical Imaging and Vision}, 57\penalty0
  (1):\penalty0 1--25, 2017.
\newblock \doi{10.1007/s10851-016-0662-8}.

\bibitem[Deng et~al.(2009)Deng, Dong, Socher, Li, Li, and
  Fei-Fei]{deng2009imagenet}
Jia Deng, Wei Dong, Richard Socher, Li-Jia Li, Kai Li, and Li~Fei-Fei.
\newblock {ImageNet}: A large-scale hierarchical image database.
\newblock In \emph{IEEE Conference on Computer Vision and Pattern Recognition
  (CVPR)}, pages 248--255, 2009.
\newblock \doi{10.1109/CVPR.2009.5206848}.

\bibitem[Dhariwal and Nichol(2021)]{dhariwal2021diffusion}
Prafulla Dhariwal and Alexander Nichol.
\newblock Diffusion models beat {GANs} on image synthesis.
\newblock In \emph{Advances in Neural Information Processing Systems
  (NeurIPS)}, volume~34, pages 8780--8794, 2021.
\newblock URL
  \url{https://proceedings.neurips.cc/paper/2021/hash/49ad23d1ec9fa4bd8d77d02681df5cfa-Abstract.html}.

\bibitem[{Eastman Kodak Company}(1999)]{kodak1999}
{Eastman Kodak Company}.
\newblock {Kodak Lossless True Color Image Suite (PhotoCD PCD0992)}, 1999.
\newblock URL \url{http://r0k.us/graphics/kodak/}.

\bibitem[Feurer et~al.(2022)Feurer, Letham, Hutter, and Bakshy]{feurer2018rgpe}
Matthias Feurer, Benjamin Letham, Frank Hutter, and Eytan Bakshy.
\newblock Practical transfer learning for {B}ayesian optimization.
\newblock arXiv:1802.02219v4, 2022.
\newblock URL \url{https://arxiv.org/abs/1802.02219v4}.

\bibitem[Finn et~al.(2017)Finn, Abbeel, and Levine]{finn2017maml}
Chelsea Finn, Pieter Abbeel, and Sergey Levine.
\newblock Model-agnostic meta-learning for fast adaptation of deep networks.
\newblock In \emph{Proceedings of the International Conference on Machine
  Learning (ICML)}, volume~70 of \emph{Proceedings of Machine Learning
  Research}, pages 1126--1135. PMLR, 2017.
\newblock URL \url{https://proceedings.mlr.press/v70/finn17a.html}.

\bibitem[Goldstein and Osher(2009)]{goldstein2009split}
Tom Goldstein and Stanley Osher.
\newblock The split {Bregman} method for {L1}-regularized problems.
\newblock \emph{SIAM Journal on Imaging Sciences}, 2\penalty0 (2):\penalty0
  323--343, 2009.
\newblock \doi{10.1137/080725891}.

\bibitem[Golub et~al.(1979)Golub, Heath, and Wahba]{golub1979gcv}
Gene~H. Golub, Michael Heath, and Grace Wahba.
\newblock Generalized cross-validation as a method for choosing a good ridge
  parameter.
\newblock \emph{Technometrics}, 21\penalty0 (2):\penalty0 215--223, 1979.
\newblock \doi{10.1080/00401706.1979.10489751}.

\bibitem[Ha et~al.(2017)Ha, Dai, and Le]{ha2016hypernetworks}
David Ha, Andrew Dai, and Quoc~V. Le.
\newblock {HyperNetworks}.
\newblock In \emph{International Conference on Learning Representations
  (ICLR)}, 2017.
\newblock URL \url{https://openreview.net/forum?id=rkpACe1lx}.

\bibitem[Hansen(1992)]{hansen1992lcurve}
Per~Christian Hansen.
\newblock Analysis of discrete ill-posed problems by means of the {L}-curve.
\newblock \emph{SIAM Review}, 34\penalty0 (4):\penalty0 561--580, 1992.
\newblock \doi{10.1137/1034115}.

\bibitem[He et~al.(2016)He, Zhang, Ren, and Sun]{he2016resnet}
Kaiming He, Xiangyu Zhang, Shaoqing Ren, and Jian Sun.
\newblock Deep residual learning for image recognition.
\newblock In \emph{Proceedings of the IEEE Conference on Computer Vision and
  Pattern Recognition (CVPR)}, pages 770--778, 2016.
\newblock \doi{10.1109/CVPR.2016.90}.
\newblock URL
  \url{https://openaccess.thecvf.com/content_cvpr_2016/html/He_Deep_Residual_Learning_CVPR_2016_paper.html}.

\bibitem[Howard and Ruder(2018)]{howard2018ulmfit}
Jeremy Howard and Sebastian Ruder.
\newblock Universal language model fine-tuning for text classification.
\newblock In \emph{Proceedings of the 56th Annual Meeting of the Association
  for Computational Linguistics (Volume 1: Long Papers)}, pages 328--339, 2018.
\newblock \doi{10.18653/v1/P18-1031}.
\newblock URL \url{https://aclanthology.org/P18-1031/}.

\bibitem[Huang et~al.(2015)Huang, Singh, and Ahuja]{huang2015urban100}
Jia-Bin Huang, Abhishek Singh, and Narendra Ahuja.
\newblock Single image super-resolution from transformed self-exemplars.
\newblock In \emph{IEEE Conference on Computer Vision and Pattern Recognition
  (CVPR)}, pages 5197--5206, 2015.
\newblock \doi{10.1109/cvpr.2015.7299156}.

\bibitem[J{\o}rgensen et~al.(2021)J{\o}rgensen, Ametova, Burca, Fardell,
  Papoutsellis, Pasca, Thielemans, Turner, Warr, Lionheart, and
  Withers]{jorgensen2021cil}
Jakob~Sauer J{\o}rgensen, Evelina Ametova, Genoveva Burca, Gemma Fardell,
  Evangelos Papoutsellis, Edoardo Pasca, Kris Thielemans, Martin Turner, Ryan
  Warr, William R.~B. Lionheart, and Philip~J. Withers.
\newblock {Core Imaging Library}---part {I}: a versatile {Python} framework for
  tomographic imaging.
\newblock \emph{Philosophical Transactions of the Royal Society A},
  379\penalty0 (2204):\penalty0 20200192, 2021.
\newblock \doi{10.1098/rsta.2020.0192}.

\bibitem[Karras et~al.(2019)Karras, Laine, and Aila]{karras2019stylegan}
Tero Karras, Samuli Laine, and Timo Aila.
\newblock A style-based generator architecture for generative adversarial
  networks.
\newblock In \emph{Proceedings of the IEEE/CVF Conference on Computer Vision
  and Pattern Recognition (CVPR)}, pages 4401--4410, 2019.
\newblock \doi{10.1109/CVPR.2019.00453}.
\newblock URL
  \url{https://openaccess.thecvf.com/content_CVPR_2019/html/Karras_A_Style-Based_Generator_Architecture_for_Generative_Adversarial_Networks_CVPR_2019_paper.html}.

\bibitem[Kazantsev et~al.(2019)Kazantsev, Pasca, Turner, and
  Withers]{kazantsev2019ccpi}
Daniil Kazantsev, Edoardo Pasca, Martin~J. Turner, and Philip~J. Withers.
\newblock {CCPi-Regularisation} toolkit for computed tomographic image
  reconstruction with proximal splitting algorithms.
\newblock \emph{SoftwareX}, 9:\penalty0 317--323, 2019.
\newblock \doi{10.1016/j.softx.2019.04.003}.

\bibitem[Kendall et~al.(2018)Kendall, Gal, and Cipolla]{kendall2017multitask}
Alex Kendall, Yarin Gal, and Roberto Cipolla.
\newblock Multi-task learning using uncertainty to weigh losses for scene
  geometry and semantics.
\newblock In \emph{Proceedings of the IEEE/CVF Conference on Computer Vision
  and Pattern Recognition (CVPR)}, pages 7482--7491, 2018.
\newblock \doi{10.1109/CVPR.2018.00781}.

\bibitem[Loshchilov and Hutter(2019)]{loshchilov2017adamw}
Ilya Loshchilov and Frank Hutter.
\newblock Decoupled weight decay regularization.
\newblock In \emph{International Conference on Learning Representations
  (ICLR)}, 2019.
\newblock URL \url{https://openreview.net/forum?id=Bkg6RiCqY7}.

\bibitem[Ma et~al.(2017)Ma, Duanmu, Wu, Wang, Yong, Li, and
  Zhang]{ma2017waterloo}
Kede Ma, Zhengfang Duanmu, Qingbo Wu, Zhou Wang, Hongwei Yong, Hongliang Li,
  and Lei Zhang.
\newblock {Waterloo Exploration Database}: New challenges for image quality
  assessment models.
\newblock \emph{IEEE Transactions on Image Processing}, 26\penalty0
  (2):\penalty0 1004--1016, 2017.
\newblock \doi{10.1109/TIP.2016.2631888}.

\bibitem[{Maboudi Afkham} et~al.(2021){Maboudi Afkham}, Chung, and
  Chung]{afkham2021learning}
Babak {Maboudi Afkham}, Julianne Chung, and Matthias Chung.
\newblock Learning regularization parameters of inverse problems via deep
  neural networks.
\newblock \emph{Inverse Problems}, 37\penalty0 (10):\penalty0 105017, 2021.
\newblock \doi{10.1088/1361-6420/ac245d}.

\bibitem[M{\"a}kinen et~al.(2020)M{\"a}kinen, Azzari, and
  Foi]{makinen2020collaborative}
Ymir M{\"a}kinen, Lucio Azzari, and Alessandro Foi.
\newblock Collaborative filtering of correlated noise: Exact transform-domain
  variance for improved shrinkage and patch matching.
\newblock \emph{IEEE Transactions on Image Processing}, 29:\penalty0
  8339--8354, 2020.
\newblock \doi{10.1109/TIP.2020.3014721}.

\bibitem[Martin et~al.(2001)Martin, Fowlkes, Tal, and Malik]{martin2001bsds}
David~R. Martin, Charless~C. Fowlkes, Doron Tal, and Jitendra Malik.
\newblock A database of human segmented natural images and its application to
  evaluating segmentation algorithms and measuring ecological statistics.
\newblock In \emph{Proceedings of the IEEE International Conference on Computer
  Vision (ICCV)}, volume~2, pages 416--423, 2001.
\newblock \doi{10.1109/ICCV.2001.937655}.

\bibitem[Nichol et~al.(2018)Nichol, Achiam, and Schulman]{nichol2018reptile}
Alex Nichol, Joshua Achiam, and John Schulman.
\newblock On first-order meta-learning algorithms.
\newblock arXiv:1803.02999, 2018.
\newblock URL \url{https://arxiv.org/abs/1803.02999}.

\bibitem[Ramani et~al.(2008)Ramani, Blu, and Unser]{ramani2008sure}
S.~Ramani, T.~Blu, and M.~Unser.
\newblock Monte-{C}arlo {SURE}: A black-box optimization of regularization
  parameters for general denoising algorithms.
\newblock \emph{IEEE Transactions on Image Processing}, 17\penalty0
  (9):\penalty0 1540--1554, 2008.
\newblock \doi{10.1109/TIP.2008.2001404}.

\bibitem[Rudin et~al.(1992)Rudin, Osher, and Fatemi]{rudin1992tv}
Leonid~I. Rudin, Stanley Osher, and Emad Fatemi.
\newblock Nonlinear total variation based noise removal algorithms.
\newblock \emph{Physica D: Nonlinear Phenomena}, 60\penalty0 (1--4):\penalty0
  259--268, 1992.
\newblock \doi{10.1016/0167-2789(92)90242-f}.

\bibitem[Vu et~al.(2025)Vu, Kofler, and Papafitsoros]{vu2025tgv}
Thanh~Trung Vu, Andreas Kofler, and Kostas Papafitsoros.
\newblock Deep unrolling for learning optimal spatially varying regularisation
  parameters for total generalised variation.
\newblock In \emph{Scale Space and Variational Methods in Computer Vision},
  volume 15667 of \emph{Lecture Notes in Computer Science}, pages 282--294.
  Springer Nature Switzerland, 2025.
\newblock \doi{10.1007/978-3-031-92366-1_22}.
\newblock URL \url{https://arxiv.org/abs/2502.16532}.

\bibitem[Wang et~al.(2022)Wang, Dalca, and Sabuncu]{wang2022hyperrecon}
Alan~Q. Wang, Adrian~V. Dalca, and Mert~R. Sabuncu.
\newblock Computing multiple image reconstructions with a single hypernetwork.
\newblock \emph{Machine Learning for Biomedical Imaging}, 2022\penalty0 (June
  2022 issue):\penalty0 1--25, 2022.
\newblock \doi{10.59275/j.melba.2022-e5ec}.
\newblock URL \url{https://melba-journal.org/2022:017}.

\bibitem[Wang et~al.(2004)Wang, Bovik, Sheikh, and Simoncelli]{wang2004ssim}
Zhou Wang, Alan~C. Bovik, Hamid~R. Sheikh, and Eero~P. Simoncelli.
\newblock Image quality assessment: From error visibility to structural
  similarity.
\newblock \emph{IEEE Transactions on Image Processing}, 13\penalty0
  (4):\penalty0 600--612, 2004.
\newblock \doi{10.1109/TIP.2003.819861}.

\bibitem[Wei et~al.(2020)Wei, Aviles-Rivero, Liang, Fu, Sch{\"o}nlieb, and
  Huang]{wei2020tfpnp}
Kaixuan Wei, Angelica~I. Aviles-Rivero, Jingwei Liang, Ying Fu, Carola-Bibiane
  Sch{\"o}nlieb, and Hua Huang.
\newblock Tuning-free plug-and-play proximal algorithm for inverse imaging
  problems.
\newblock In \emph{Proceedings of the International Conference on Machine
  Learning (ICML)}, volume 119 of \emph{Proceedings of Machine Learning
  Research}, pages 10158--10169. PMLR, 2020.
\newblock URL \url{https://proceedings.mlr.press/v119/wei20b.html}.

\bibitem[Wistuba and Grabocka(2021)]{wistuba2021fsbo}
Martin Wistuba and Josif Grabocka.
\newblock Few-shot {B}ayesian optimization with deep kernel surrogates.
\newblock In \emph{International Conference on Learning Representations
  (ICLR)}, 2021.
\newblock URL \url{https://openreview.net/forum?id=bJxgv5C3sYc}.

\bibitem[Woo et~al.(2023)Woo, Debnath, Hu, Chen, Liu, Kweon, and
  Xie]{woo2023convnextv2}
Sanghyun Woo, Shoubhik Debnath, Ronghang Hu, Xinlei Chen, Zhuang Liu, In~So
  Kweon, and Saining Xie.
\newblock {ConvNeXt V2}: Co-designing and scaling {ConvNets} with masked
  autoencoders.
\newblock In \emph{Proceedings of the IEEE/CVF Conference on Computer Vision
  and Pattern Recognition (CVPR)}, pages 16133--16142, 2023.
\newblock \doi{10.1109/cvpr52729.2023.01548}.

\bibitem[Zhang et~al.(2018)Zhang, Isola, Efros, Shechtman, and
  Wang]{zhang2018lpips}
Richard Zhang, Phillip Isola, Alexei~A. Efros, Eli Shechtman, and Oliver Wang.
\newblock The unreasonable effectiveness of deep features as a perceptual
  metric.
\newblock In \emph{Proceedings of the IEEE Conference on Computer Vision and
  Pattern Recognition (CVPR)}, pages 586--595, 2018.
\newblock \doi{10.1109/CVPR.2018.00068}.
\newblock URL
  \url{https://openaccess.thecvf.com/content_cvpr_2018/html/Zhang_The_Unreasonable_Effectiveness_CVPR_2018_paper.html}.

\bibitem[Zhu et~al.(2023)Zhu, Zhang, Liang, Cao, Wen, Timofte, and
  Van~Gool]{zhu2023denoising}
Yuanzhi Zhu, Kai Zhang, Jingyun Liang, Jiezhang Cao, Bihan Wen, Radu Timofte,
  and Luc Van~Gool.
\newblock Denoising diffusion models for plug-and-play image restoration.
\newblock In \emph{IEEE/CVF Conference on Computer Vision and Pattern
  Recognition Workshops (CVPRW)}, pages 1219--1229, 2023.
\newblock \doi{10.1109/cvprw59228.2023.00129}.

\end{thebibliography}
\endgroup

\clearpage
\appendix

\noindent
The appendices are organized as follows: implementation and experimental setup (Appendix~\ref{app:impl_details}), transfer protocols and additional results (Appendix~\ref{app:additional_generalization}), architecture ablations (Appendix~\ref{sec:arch_validation}), computational cost and classical comparisons (Appendix~\ref{app:computational_cost}), and predictor transfer: definitions and proof (Appendix~\ref{app:transfer_theory}).

\section{Implementation and experimental setup}
\label{app:impl_details}

\subsection{Source configurations and noise generation}

Table~\ref{tab:source_bundles} lists the source settings: $12$ STL10 $96{\times}96$ configurations and $1$ WED $256{\times}256$ configuration. These settings cover Gaussian, Poisson, Poisson--Gaussian, and black-and-white (bw) or channel-wise (rgb) impulse noise, alone or mixed with Gaussian noise.

\begin{table}[!htbp]
\centering
\setlength{\tabcolsep}{4pt}
\caption{Source settings. ``bw'' and ``rgb'' denote black-and-white and channel-wise impulse noise; NLL denotes the likelihood-based data term used for Poisson or Poisson--Gaussian observations.}
\label{tab:source_bundles}
\begin{tabular}{clp{0.31\linewidth}ll}
\toprule
\# & Dataset/res. & Noise & Data term & Prior \\
\midrule
1 & STL10 $96{\times}96$ & bw impulse $+$ Gaussian & Huber & TV \\
2 & STL10 $96{\times}96$ & bw impulse & Huber & TGV \\
3 & STL10 $96{\times}96$ & bw impulse & Huber & TV \\
4 & STL10 $96{\times}96$ & Gaussian & Huber & TGV \\
5 & STL10 $96{\times}96$ & Gaussian & L2 & TGV \\
6 & STL10 $96{\times}96$ & Gaussian & L2 & TV \\
7 & WED $256{\times}256$ crops & Gaussian & L2 & TV \\
8 & STL10 $96{\times}96$ & Gaussian & Huber & TV \\
9 & STL10 $96{\times}96$ & Poisson--Gaussian & NLL & TGV \\
10 & STL10 $96{\times}96$ & Poisson--Gaussian & NLL & TV \\
11 & STL10 $96{\times}96$ & Poisson & NLL & TV \\
12 & STL10 $96{\times}96$ & rgb impulse $+$ Gaussian & Huber & TGV \\
13 & STL10 $96{\times}96$ & rgb impulse & Huber & TGV \\
\bottomrule
\end{tabular}
\end{table}

In the synthetic-noise experiments, each noise realization uses a deterministic seed and is kept fixed across the oracle sweep, training, and evaluation. In Table~\ref{tab:source_bundles}, source rows $6$, $8$, $10$, and $11$ each contain four noise realizations per clean image, giving $20\,000$ records from $5\,000$ images. Row $9$ contains $100$ records, and the other STL10 sources contain $5\,000$ each. The WED source (row $7$) contains $4\,344$ records. Records are oracle-labeled noisy observations; different noise realizations of one clean image count separately. Noise is sampled as follows:
\begin{itemize}
    \item \textbf{Gaussian noise (DiffPIR few-shot target on WED $256{\times}256$ crops, Gaussian source training sets, and classical-comparison test sets):} per-observation standard deviation $\sigma\sim\mathrm{Uniform}(5/255, 50/255)$, applied additively per pixel.
    \item \textbf{Black-and-white impulse:} salt-and-pepper noise sharing locations and values across channels. Each observation uses a corruption probability drawn from $\mathrm{Uniform}(0.05, 0.30)$, with salt and pepper equally likely.
    \item \textbf{RGB impulse:} salt-and-pepper noise applied independently to each channel. Each observation uses one corruption probability drawn from $\mathrm{Uniform}(0.05, 0.30)$ for all channels, with salt and pepper equally likely.
    \item \textbf{Poisson and Poisson--Gaussian (NLL source training sets):} per-observation photon scale is sampled as $p_{\max}\sim\mathrm{Uniform}(2,128)$ with $\alpha=1/p_{\max}$ in the Poisson observation model. Pure Poisson uses read noise $\sigma_{\mathrm{read}}=0$, while Poisson--Gaussian samples $\sigma_{\mathrm{read}}\sim\mathrm{Uniform}(0.5/255,16/255)$; the background offset is fixed to $0$.
    \item \textbf{Mixed-noise configurations} (Huber-TV bw impulse $+$ Gaussian, Huber-TGV rgb impulse $+$ Gaussian): the impulse corruption probability is sampled from the corresponding single-noise range above, Gaussian $\sigma\sim\mathrm{Uniform}(5/255,30/255)$, and the two noises are applied in sequence (impulse, then Gaussian).
\end{itemize}

\FloatBarrier

\subsection{Denoising solvers}

The following fixed denoisers are used for oracle-label generation and test-time reconstruction.
\begin{itemize}
    \item \textbf{L2-TV denoising:} \texttt{skimage.restoration.\allowbreak denoise\_tv\_bregman} implements split Bregman TV~\citep{goldstein2009split} with isotropic TV, per-channel weight $1/(2\lambda)$, and the default stopping settings (\texttt{max\_num\_iter}$=100$, \texttt{eps}$=10^{-3}$; whichever is reached first).
    \item \textbf{Huber-TV denoising:} an alternating direction method of multipliers solver~\citep{boyd2011admm} with a Huber proximal step, $100$ iterations, $\rho_{\mathrm{TV}}{=}2.0$, and $\rho_{\mathrm{data}}{=}5.0$. It tunes TV weight $\lambda$ and Huber threshold $\delta$.
    \item \textbf{L2/NLL-TGV denoising:} second-order TGV with $(\lambda,\gamma)$ passed as the CCPi~\citep{kazantsev2019ccpi} weight \texttt{alpha} and ratio parameter \texttt{gamma}. The \href{https://doi.org/10.5281/zenodo.4746198}{CIL} wrapper~\citep{jorgensen2021cil} uses the fast iterative shrinkage-thresholding algorithm (FISTA) for at most $200$ outer iterations with relative-update tolerance $10^{-4}$ (after at least $5$ iterations); each CCPi TGV proximal call uses at most $100$ iterations.
    \item \textbf{NLL-TV denoising:} CIL/FISTA with TV weight $\lambda$ and the CCPi \texttt{FGP\_TV} proximal operator for isotropic TV: at most $200$ outer iterations, objective-value stopping threshold $10^{-4}$, and at most $100$ iterations per proximal call.
    \item \textbf{Huber-TGV denoising:} the same CIL/FISTA and CCPi TGV settings with a Huber data term. It tunes $(\lambda,\gamma,\delta)$, where $\delta$ is the Huber threshold.
\end{itemize}
CIL solves each channel separately, clipping initial and final images to $[0,1]$. FGP\_TV enforces nonnegativity; TGV proximal outputs are clipped to $[0,1]$. All five denoisers run deterministically on CPU. DiffPIR uses the public $100$-iteration sampler (\S\ref{sec:exp_cross_paradigm}; Appendix~\ref{app:diffpir_transfer}).

\subsection{Oracle labels and training objective}

All experiments use oracle labels produced by hierarchical grid searches. Except for the metric-specific SSIM and LPIPS study, the target hyperparameter vector maximizes PSNR for each image--configuration pair; that study instead optimizes its named metric. HyperDn fits encoded predictions $\hat{\bm q}$ to the encoded oracle labels $\widetilde{\bm q}$ over the active slots, as described in Section~\ref{sec:pooled_supervision}. For a configuration $\chi$, let $\mathcal{A}_{\chi}$ be the set of slots that are real hyperparameters rather than padding. For a minibatch $\mathcal B$, let $\mathcal B_\chi$ contain its examples from configuration $\chi$, and let $Z_{\mathcal B}=\sum_\chi|\mathcal B_\chi||\mathcal A_\chi|$ count its active scalar targets. The unweighted minibatch loss is
\begin{equation}
\label{eq:active_slot_mse}
\mathcal{L}_{\mathrm{slot}}(\Theta;\mathcal B)
=
\frac{1}{Z_{\mathcal B}}
\sum_\chi
\sum_{u\in\mathcal B_\chi}
\sum_{j\in\mathcal{A}_{\chi}}
\bigl(\hat q_{\chi,u,j}-\widetilde q_{\chi,u,j}\bigr)^2.
\end{equation}
The zero-shot mixed-noise runs use this plain active-slot MSE. The main joint pretraining run and HyperDn/ImageNet init finetuning on DiffPIR, NLM/CBM3D, and SIDD use learned per-role loss weights following \citet{kendall2017multitask}. Let $s_\rho=\log\sigma_\rho^2$ be the role log-variance, clipped to $[-10,10]$ when computing the loss, and let $\rho_{\chi,j}$ be the role of active slot $j$ under configuration $\chi$. The weighted minibatch loss is
\begin{equation}
\label{eq:role_weighted_loss}
\mathcal{L}_{\mathrm{role}}(\Theta,\bm{s};\mathcal B)
=
\frac{1}{Z_{\mathcal B}}
\sum_\chi
\sum_{u\in\mathcal B_\chi}
\sum_{j\in\mathcal{A}_{\chi}}
\left[
\frac{1}{2}\exp(-s_{\rho_{\chi,j}})
\bigl(\hat q_{\chi,u,j}-\widetilde q_{\chi,u,j}\bigr)^2
+ \frac{1}{2}s_{\rho_{\chi,j}}
\right].
\end{equation}
This loss matches the Gaussian negative log-likelihood form up to constants independent of $\Theta$ and $\bm{s}$. For HyperDn, DiffPIR finetuning starts from the log-variance values learned during source pretraining.

\subsection{Predictor architecture and source pretraining}

The encoder uses the stem and stages $0$--$1$ of the ConvNeXtV2-base \texttt{fcmae\_ft\_in22k\_in1k} checkpoint~\citep{woo2023convnextv2}. The denoiser-calibrated readout has $L_{\max}{=}3$ slots to accommodate Huber-TGV's $(\lambda,\gamma,\delta)$. Role and scale embeddings have $32$ dimensions; metadata embeddings have $16$. The two-layer trunk MLP uses GELU, LayerNorm, dropout $0.1$, and a $512$-dimensional output $\bm{h}$. Each role has a learned log-variance $\log\sigma^2$ for loss weighting.

We use AdamW~\citep{loshchilov2017adamw} for $80$ epochs at batch size $128$, head learning rate $2{\times}10^{-4}$, and layer-wise learning-rate decay (LLRD; \citealp{howard2018ulmfit}) of $0.8$. We use weight decay $10^{-5}$, a cosine schedule to $10^{-6}$ after $10\%$ linear warmup, gradient clipping at $1.0$, and mixed precision. Augmentations are random horizontal/vertical flips and $90^\circ/270^\circ$ rotations; we apply no condition-embedding dropout. We randomly hold out $5\%$ of the combined source records and select the checkpoint with the lowest weighted validation loss. Each batch contains one source so image sizes match; its loss averages active scalar targets. We shuffle batches across sources.

\subsection{Statistical reporting}

We compute standard deviations across seeds, after averaging each seed's metric over the test images. DiffPIR (including metric-specific transfer, ResNet50, and CBSD68/Urban100), SIDD, and FFHQ use sample standard deviation (divisor $n-1$ for $n$ seeds). Mixed-noise, cross-resolution, NLM/CBM3D, architecture-ablation, and classical-comparison results use population standard deviation (divisor $n$). Fixed controls derived from training-label subsets use the same convention as their learned counterparts. Captions identify three-seed exceptions. Oracles use fixed evaluation images and deterministic searches over finite candidate sets.

\subsection{Source and target settings}

\begin{table}[!htbp]
\centering
\setlength{\tabcolsep}{3pt}
\caption{Sources and targets for transfer experiments (continued in Table~\ref{tab:transfer_protocol_map_additional}). HyperDn receives the degraded image and target denoiser identity. ``Observation metadata'' denotes numeric acquisition or noise parameters; noise-type information is represented by categorical codes. $N$ counts target oracle-labeled images used for predictor fitting; DiffPIR's shared calibration uses additional labels (Appendix~\ref{app:diffpir_transfer}). The architecture ablations reuse the Huber-TV mixed-noise protocol and change only the tested module.}
\label{tab:transfer_protocol_map}
\begin{tabular}{@{}>{\raggedright\arraybackslash}p{.14\textwidth}>{\raggedright\arraybackslash}p{.25\textwidth}>{\raggedright\arraybackslash}p{.24\textwidth}>{\raggedright\arraybackslash}p{.09\textwidth}>{\raggedright\arraybackslash}p{.20\textwidth}@{}}
\toprule
Experiment & HyperDn source supervision & Target / evaluation & Target labels $N$ & Test-time information beyond the degraded image \\
\midrule
DiffPIR/\allowbreak WED (PSNR) & TV/TGV source checkpoint (Table~\ref{tab:source_bundles}) & WED $256{\times}256$; DiffPIR with ImageNet prior and Gaussian noise & $1$, $2$, $4$, $16$, $32$, $64$, $128$ & Gaussian category; $\zeta=1$; no observation metadata \\
DiffPIR/\allowbreak WED (SSIM, LPIPS) & Fresh two-source WED TV/TGV checkpoint for each named objective & WED $256{\times}256$; DiffPIR with ImageNet prior and Gaussian noise & $1$, $2$, $4$, $16$, $32$, $64$, $128$ & Gaussian category; $\zeta=1$; no observation metadata \\
\midrule
Mixed noise, Huber-TV & Two single-noise STL10 sets: Huber-TV/bw impulse and Huber-TV/Gaussian & STL10 test; bw impulse $+$ Gaussian & $0$ & Known composition code; no observation metadata; all-zero-code ablation also tested \\
Mixed noise, Huber-TGV & Two single-noise STL10 sets: Huber-TGV/rgb impulse and Huber-TGV/Gaussian & STL10 test; rgb impulse $+$ Gaussian & $0$ & Known composition code; no observation metadata; all-zero-code ablation also tested \\
\midrule
Cross resolution & TV/TGV source checkpoint (Table~\ref{tab:source_bundles}) & Kodak24: Huber-TV/bw impulse and L2-TGV/Gaussian; CBSD68, Kodak24, Urban100: L2-TV/Gaussian & $0$ & Denoiser/noise code and input size; no observation metadata \\
\bottomrule
\end{tabular}
\end{table}

\clearpage
\begin{table}[!htbp]
\centering
\setlength{\tabcolsep}{3pt}
\caption{Source supervision and target evaluation for additional transfer experiments and classical comparisons. Definitions follow Table~\ref{tab:transfer_protocol_map}.}
\label{tab:transfer_protocol_map_additional}
\begin{tabular}{@{}>{\raggedright\arraybackslash}p{.14\textwidth}>{\raggedright\arraybackslash}p{.25\textwidth}>{\raggedright\arraybackslash}p{.24\textwidth}>{\raggedright\arraybackslash}p{.09\textwidth}>{\raggedright\arraybackslash}p{.20\textwidth}@{}}
\toprule
Experiment & HyperDn source supervision & Target / evaluation & Target labels $N$ & Test-time information beyond the degraded image \\
\midrule
NLM / CBM3D & TV/TGV source checkpoint (Table~\ref{tab:source_bundles}); neither target denoiser nor its scalar appears in source training & WED $256{\times}256$; NLM $h$ or CBM3D $\sigma_{\rm psd}$ & $1$, $2$, $4$, $16$, $32$, $64$, $128$ & Target denoiser and \texttt{UNKNOWN} parameter role; no observation metadata \\
DiffPIR/\allowbreak CBSD68, Urban100 & Seed-matched TV/TGV source checkpoints (Table~\ref{tab:source_bundles}); no WED DiffPIR labels for predictor fitting & Dataset-specific $256{\times}256$ crops; DiffPIR with ImageNet prior & $1$, $2$, $4$, $16$ & Gaussian category; $\zeta=1$; no observation metadata \\
SIDD sRGB \texttt{UNKNOWN} & Fresh $10$-source checkpoint: Table~\ref{tab:source_bundles} excluding Poisson-family rows 9--11 & SIDD Medium training crops; official Validation sRGB blocks; Huber-TV & $2$ & \texttt{UNKNOWN} noise code; zero observation metadata \\
DiffPIR/\allowbreak FFHQ prior & Source checkpoint, then $128$ WED DiffPIR labels generated with the ImageNet prior & FFHQ faces; DiffPIR with an FFHQ-trained prior & $0$ FFHQ & Gaussian category; $\zeta=1$; no observation metadata \\
Classical/\allowbreak SURE comparison & TV/TGV source checkpoint (Table~\ref{tab:source_bundles}) & L2-TV/Gaussian on CBSD68, Kodak24, and Urban100 & $0$ & HyperDn: Gaussian category, no $\sigma$ or observation metadata; SURE: true $\sigma$ \\
\bottomrule
\end{tabular}
\end{table}

\clearpage
\section{Transfer protocols and additional results}
\label{app:additional_generalization}

\subsection{Few-shot transfer to DiffPIR}
\label{app:diffpir_transfer}

\paragraph{DiffPIR few-shot finetuning.}
HyperDn and ImageNet init use the same DiffPIR finetuning schedule: $150$ epochs, head learning rate $\mathrm{LR}_{\mathrm{head}}=10^{-4}$, LLRD of $0.1$, $30\%$ linear warmup, batch size at most $8$ (or the target-label count when smaller), AdamW with weight decay $10^{-5}$ and gradient clipping at $1.0$, no mixed precision, and the role-weighted loss in Eq.~(\ref{eq:role_weighted_loss}). We use all target labels for finetuning and select the last-epoch checkpoint.

HyperDn initializes from the \emph{full} source-pretrained checkpoint (backbone, head, embeddings, and weighting parameters). The denoiser-dependent scale and offset tables in Eq.~(\ref{eq:slot-readout}) are initialized at $1$ and $0$ before source pretraining. Their DiffPIR-specific rows receive no source-task gradients and are loaded from the source checkpoint for target finetuning. ImageNet init loads the ImageNet-pretrained ConvNeXtV2 backbone and uses a randomly initialized head (seeded per run).

\paragraph{Afkham baseline.}
Afkham trains from random initialization for each label budget and seed: $500$ epochs for $N\le2$ and $100$ otherwise. It uses AdamW with learning rate and weight decay both $10^{-3}$, batch size $\min(N,8)$, mixed precision, and a cosine schedule after one warmup epoch. It minimizes unweighted prediction MSE on all $N$ labels and uses the final checkpoint. Following the convolutional design of \citet{afkham2021learning}, adapted from the original single-channel deblurring setting to RGB DiffPIR denoising, this target-only neural network predicts regularization parameters for one inverse-problem configuration. It consists of three convolutional blocks (Conv $5{\times}5$, padding 2 \,$\to$\, BatchNorm \,$\to$\, ReLU \,$\to$\, MaxPool $2{\times}2$, with channel widths $3{\to}16{\to}32{\to}64$), an adaptive average pool to $4{\times}4$, and an MLP head with input width $1024$, hidden widths $256{\to}128{\to}64$, and ReLU between hidden layers, followed by a linear output for the active DiffPIR target; this is approximately $369$\,K parameters in total.

\paragraph{ImageNet-pretrained ResNet50 control.}
We also fully finetune a standard ImageNet-V1-pretrained ResNet50~\citep{he2016resnet} on the same DiffPIR training subsets for each seed and evaluate it on the same $50$ test images. Every trainable layer uses the same peak learning rate $10^{-5}$, with no layer-wise decay; otherwise it follows the $150$-epoch target schedule above. This compares HyperDn with ResNet50 at the same target-label budgets.

\paragraph{DiffPIR reconstruction and evaluation.}
We predict only $\log_{10}\lambda_{\mathrm{eff}}$, where $\lambda_{\mathrm{eff}}=\lambda_{\mathrm{cache}}\sigma^2$ converts the cached DiffPIR search parameter to the solver-interface coefficient. This conversion uses the training observation's noise variance; no noise-level input is needed for prediction. We reconstruct with $100$ DiffPIR iterations using the ImageNet-trained guided-diffusion prior~\citep{dhariwal2021diffusion}; every method fixes $\zeta=1$, selected because all $4{,}344$ WED training images chose it in a preliminary joint $(\lambda_{\mathrm{cache}},\zeta)$ sweep. The reported few-shot budgets count image-specific $\lambda_{\mathrm{eff}}$ labels used for predictor fitting, conditional on this shared calibration. The preliminary sweep uses additional target-side supervision and is excluded from those fitting budgets. Evaluation uses the same deterministic first $50$ images of the $400$-image validation split, identical degraded inputs, and float-$[0,1]$ PSNR without border cropping; the oracle PSNR is $31.13$\,dB.

\paragraph{Complete budget results.}
Table~\ref{tab:cross_paradigm_diffpir} gives the exact values behind the few-shot transfer curve in Section~\ref{sec:exp_cross_paradigm} and includes the ResNet50 control.

\begin{table}[!htbp]
\centering
\setlength{\tabcolsep}{6pt}
\caption{DiffPIR PSNR (dB) $\uparrow$ across all target-label budgets; oracle: $31.13$\,dB. Best mean per budget in \textbf{bold}.}
\label{tab:cross_paradigm_diffpir}
\begin{tabular}{@{}rrrrr@{}}
\toprule
$N$ & Afkham & ImageNet init & ResNet50 & HyperDn \\
\midrule
1 & $23.06\!\pm\!1.66$ & $24.98\!\pm\!2.32$ & $27.31\!\pm\!.94$ & $\mathbf{29.37\!\pm\!.77}$ \\
2 & $27.26\!\pm\!1.79$ & $27.73\!\pm\!1.40$ & $27.96\!\pm\!.77$ & $\mathbf{30.23\!\pm\!.56}$ \\
4 & $28.23\!\pm\!2.00$ & $28.69\!\pm\!.89$ & $27.75\!\pm\!.99$ & $\mathbf{30.62\!\pm\!.23}$ \\
16 & $27.84\!\pm\!2.46$ & $29.17\!\pm\!.95$ & $28.70\!\pm\!.40$ & $\mathbf{30.84\!\pm\!.12}$ \\
32 & $29.64\!\pm\!.53$ & $30.53\!\pm\!.29$ & $29.63\!\pm\!.36$ & $\mathbf{31.03\!\pm\!.01}$ \\
64 & $30.28\!\pm\!.40$ & $30.90\!\pm\!.09$ & $30.44\!\pm\!.20$ & $\mathbf{31.09\!\pm\!.01}$ \\
128 & $30.81\!\pm\!.08$ & $30.94\!\pm\!.14$ & $30.81\!\pm\!.08$ & $\mathbf{31.10\!\pm\!.01}$ \\
\bottomrule
\end{tabular}
\end{table}

\clearpage

\subsection{Zero-shot mixed-noise transfer}
\label{app:mixed_noise_transfer}

For each mixed-noise target, we train a fresh predictor using the two corresponding single-noise source sets from the STL10 \emph{train} split ($5\,000$ clean images at $96{\times}96$; source record counts are given in Appendix~\ref{app:impl_details}): Huber--TV/bw impulse with Huber--TV/Gaussian, or Huber--TGV/rgb impulse with Huber--TGV/Gaussian. Mixed-noise target evaluation in Table~\ref{tab:exp_a_composition} uses $1\,000$ images from the STL10 \emph{test} split at $96{\times}96$, with no overlap with the training images; no mixture label enters training or model selection. We follow the source-pretraining optimizer, schedule, and augmentations in Appendix~\ref{app:impl_details}, with $80$ epochs, batch size $256$, and equally weighted MSE over the active parameter dimensions.

The multi-hot noise-attribute code has six slots: (unknown, has-Gaussian, has-Poisson, has-impulse, impulse-is-bw, impulse-is-rgb). Thus bw impulse $+$ Gaussian encodes to $[0,1,0,1,1,0]$, and rgb impulse $+$ Gaussian encodes to $[0,1,0,1,0,1]$. Neither target receives observation metadata. Replacing all six categorical-code entries with zeros at inference changes the five-seed mean PSNR by $-0.0074$ and $-0.0077$\,dB on the bw and rgb mixture targets, respectively, relative to using the mixture codes above.

\subsection{Cross-resolution transfer}
\label{app:cross_resolution_transfer}

For cross-resolution evaluation on Kodak24 ($512{\times}768$), L2-TGV uses the Gaussian noise range of its STL10 source, and Huber-TV uses the black-and-white impulse corruption-probability range of its STL10 source. For L2-TV, source supervision includes Gaussian-corrupted STL10 $96{\times}96$ and WED $256{\times}256$ (Table~\ref{tab:source_bundles}). The predictor uses the joint source-pretrained checkpoint described in Appendix~\ref{app:impl_details}; results are reported in Table~\ref{tab:cross_dataset} and Figure~\ref{fig:cross_dataset}.

\subsection{Previously unseen nonlocal denoisers and parameter roles}
\label{app:nonlocal_transfer}

NLM, CBM3D, and their tuned scalars are unseen during source pretraining. Both targets use additive Gaussian noise with per-image $\sigma\sim\mathrm{Uniform}(5/255,50/255)$ and data-term identifier $d=\mathrm{L2}$. We tune NLM's filtering strength $h$ in \texttt{skimage.restoration.\allowbreak denoise\_nl\_means} and CBM3D's assumed noise standard deviation $\sigma_{\rm psd}$ in \texttt{bm3d.bm3d\_rgb}~\citep{makinen2020collaborative}, using RGB input and package defaults for the remaining denoising settings. In both cases, we use a hierarchical grid search on a logarithmic scale over $[0.003,0.30]$. We assign the scalar the \texttt{UNKNOWN} role, fine-tune on $N$ images selected by seed from a $128$-image WED $256{\times}256$ target-label pool, and evaluate on $50$ test images outside this pool. HyperDn begins from the variational source checkpoint. ImageNet init uses the identical architecture with an ImageNet-pretrained backbone and a randomly initialized head; both follow the $150$-epoch finetuning schedule in Appendix~\ref{app:diffpir_transfer}. Afkham is trained from scratch for $500$ epochs when $N\le2$ and $100$ epochs otherwise, using AdamW with learning rate and weight decay both $10^{-3}$, batch size $\min(N,8)$, and a cosine schedule after one warmup epoch. It minimizes unweighted prediction MSE and uses the final checkpoint.

Table~\ref{tab:nonlocal_transfer} reports the full results. HyperDn has the highest mean PSNR at all seven budgets despite the change in hyperparameter meaning.

\subsection{Cross-dataset transfer and real camera noise}
\label{app:dataset_noise_transfer}

Table~\ref{tab:generalization_summary} summarizes results at selected label budgets for additional datasets and real camera noise.

\begin{table}[!htbp]
\centering
\setlength{\tabcolsep}{3.5pt}
\caption{Additional dataset and real-noise targets: PSNR (dB) $\uparrow$. Controls use the listed $N$. For SIDD, the noise code is \texttt{UNKNOWN} and the observation metadata is zero. The oracle is found by per-image search over a finite candidate set. Detailed results appear below.}
\label{tab:generalization_summary}
\begin{tabular}{@{}llrrr@{}}
\toprule
Shift / target & $N$ & HyperDn & Strongest control & Oracle \\
\midrule
New dataset: DiffPIR/CBSD68 & $16$ & $\mathbf{30.05}$ & $27.30$ (Afkham) & $30.18$ \\
New dataset: DiffPIR/Urban100 & $16$ & $\mathbf{31.18}$ & $27.92$ (Afkham) & $31.41$ \\
Real noise: SIDD sRGB \texttt{UNKNOWN} & $2$ & $\mathbf{33.732}$ & $31.893$ (fixed) & $35.640$ \\
\bottomrule
\end{tabular}
\end{table}

\clearpage

\subsubsection{DiffPIR on CBSD68 and Urban100}

We repeat DiffPIR few-shot transfer on $256{\times}256$ center crops from CBSD68 and Urban100. The first $50$ images form the evaluation set; the remaining $18$ CBSD68 or $50$ Urban100 images form the target-label pool. HyperDn starts from the non-DiffPIR source checkpoint with the same seed, while Afkham starts from scratch; no WED DiffPIR oracle label is used to fit either predictor. All methods use $100$ DiffPIR iterations, the shared $\zeta=1$ setting from Appendix~\ref{app:diffpir_transfer}, and the same per-image Gaussian range as the main WED experiment.

\begin{table}[!htbp]
\centering
\setlength{\tabcolsep}{3.5pt}
\caption{Cross-dataset DiffPIR PSNR (dB) $\uparrow$ (three training seeds), evaluated on the same $50$ held-out images per dataset. Parenthetical values are oracle PSNR.}
\label{tab:diffpir_cross_dataset_full}
\begin{tabular}{@{}llrrrr@{}}
\toprule
Dataset (oracle) & Method & $N{=}1$ & $N{=}2$ & $N{=}4$ & $N{=}16$ \\
\midrule
CBSD68 ($30.18$) & HyperDn & $\mathbf{28.98\!\pm\!.52}$ & $\mathbf{29.05\!\pm\!.56}$ & $\mathbf{29.46\!\pm\!.40}$ & $\mathbf{30.05\!\pm\!.02}$ \\
 & Afkham & $23.35\!\pm\!3.16$ & $25.65\!\pm\!1.40$ & $26.78\!\pm\!.21$ & $27.30\!\pm\!.14$ \\
\midrule
Urban100 ($31.41$) & HyperDn & $\mathbf{28.91\!\pm\!2.23}$ & $\mathbf{30.39\!\pm\!.45}$ & $\mathbf{30.79\!\pm\!.08}$ & $\mathbf{31.18\!\pm\!.05}$ \\
 & Afkham & $21.92\!\pm\!.45$ & $25.56\!\pm\!2.18$ & $27.87\!\pm\!.39$ & $27.92\!\pm\!.93$ \\
\bottomrule
\end{tabular}
\end{table}

\FloatBarrier

\subsubsection{Real SIDD sRGB noise}

We test two-shot adaptation on SIDD sRGB~\citep{abdelhamed2018sidd} using Huber--TV with the categorical noise code set to \texttt{UNKNOWN} and the observation-metadata vector set to zero. For this experiment, HyperDn is pretrained on $10$ source configurations, excluding all three Poisson and Poisson--Gaussian configurations. For each seed, it is then adapted using two target labels drawn from $160$ deterministic SIDD Medium crops. HyperDn and ImageNet init follow the $150$-epoch finetuning schedule in Appendix~\ref{app:diffpir_transfer}, with the same initialization rules and final-checkpoint selection. Evaluation uses all $1{,}280$ official Validation sRGB blocks grouped into $40$ captures. The fixed control uses the geometric mean of the same two target-label parameter vectors for each seed and applies it to every validation image.

\begin{table}[!htbp]
\centering
\setlength{\tabcolsep}{3pt}
\caption{SIDD sRGB PSNR with two target training labels. The noisy input is deterministic.}
\label{tab:sidd_unknown}
\begin{tabular}{@{}lrrrr@{}}
\toprule
Metric & Noisy input & Fixed (same 2 labels) & ImageNet init & HyperDn \\
\midrule
PSNR (dB) $\uparrow$ & $23.662$ & $31.893\!\pm\!2.455$ & $31.796\!\pm\!3.401$ & $\mathbf{33.732\!\pm\!.490}$ \\
\bottomrule
\end{tabular}
\end{table}

\FloatBarrier

\subsection{Zero-shot image-domain and diffusion-prior shift}
\label{app:ffhq_transfer}

Starting from the source-pretrained checkpoint, we fine-tune HyperDn using $128$ WED DiffPIR oracle labels generated with the ImageNet diffusion prior. We evaluate on the first $50$ images in DiffPIR's FFHQ validation list, resized to $256{\times}256$ with Lanczos interpolation, using the official \texttt{diffusion\_ffhq\_10m} prior. This prior is unseen by the predictor. Each image receives additive Gaussian noise with $\sigma\sim\mathrm{Uniform}(5/255,50/255)$. No FFHQ image or FFHQ-prior oracle label is used to train or adapt HyperDn. For each seed, the fixed-$\lambda$ control uses the same $128$ WED training labels as HyperDn. Each cached oracle $\lambda_{\mathrm{cache}}$ is first multiplied by its corresponding WED noise variance to obtain $\lambda_{\mathrm{eff}}$, the coefficient passed to DiffPIR. We take the geometric mean of these converted values and apply this single coefficient unchanged to every FFHQ test image. This coefficient is fixed before FFHQ evaluation, with no tuning on FFHQ or direct optimization of mean WED PSNR. Every method uses $\zeta=1$ and the FFHQ prior at evaluation.

The full results are reported in Table~\ref{tab:ffhq_prior} in the main paper.

\clearpage
\section{Architecture ablations}
\label{sec:arch_validation}

Both ablations use the Huber-TV bw impulse + Gaussian zero-shot mixed-noise setting (\S\ref{sec:exp_composition}) and the two corresponding single-noise STL10 source training sets. We vary either the prediction head or the ConvNeXtV2 stages, keeping the remaining setup fixed, to test the design choices in Section~\ref{sec:method}.

\subsection{Prediction head design}
\label{sec:exp_ablation_head}

The head ablation compares HyperDn against two alternatives with identical image, data-term, prior, and metadata embeddings: a plain MLP conditioned on the image and configuration that treats all padded output slots in the same way, and a slot-aware affine variant that scales and shifts the output using role and scale information but omits the bilinear interaction and the denoiser-dependent scale and offset. All variants share the same backbone, training schedule, and seed set; only the head differs.

Table~\ref{tab:ablation_head} reports the comparison. Parameter relative error averages $|\hat{p}_j-p_j^\star|/\max(|p_j^\star|,10^{-8})$ over active slots and then over images, using raw hyperparameter values after decoding log-scale outputs with $10^x$.

\begin{table}[!htbp]
\centering
\setlength{\tabcolsep}{5pt}
\caption{Prediction-head ablation on zero-shot mixed noise (3 seeds). PSNR and gap to oracle are in dB. All variants share the same encoder and training protocol. Parameter relative error is averaged over active slots after decoding log-scale outputs.}
\label{tab:ablation_head}
\begin{tabular}{lccc}
\toprule
Head variant & PSNR $\uparrow$ & Gap (dB) $\downarrow$ & Param. rel. err. $\downarrow$ \\
\midrule
Conditioned MLP & $23.72 \pm 0.10$ & $0.81 \pm 0.10$ & $0.255$ \\
Slot-aware affine & $24.02 \pm 0.14$ & $0.50 \pm 0.14$ & $0.188$ \\
Denoiser-calibrated (\textbf{ours}) & $\mathbf{24.29 \pm 0.09}$ & $\mathbf{0.23 \pm 0.09}$ & $\mathbf{0.160}$ \\
\bottomrule
\end{tabular}
\end{table}

Role and scale information improves prediction even without the bilinear interaction: the slot-aware affine variant improves over the plain conditioned MLP ($23.72 \to 24.02$\,dB). The full denoiser-calibrated slot readout adds another $0.27$\,dB over this intermediate variant and increases PSNR by $0.57$\,dB over the plain conditioned MLP ($23.72 \to 24.29$). It also reduces parameter relative error by $37\%$ ($0.255 \to 0.160$) and narrows the gap to oracle from $0.81$\,dB to $0.23$\,dB. These gains support the combined contribution of role and scale information, the bilinear interaction, and denoiser-specific scaling and offsets in Eq.~(\ref{eq:slot-readout}).

\FloatBarrier
\subsection{Backbone stage selection}
\label{sec:exp_ablation}

The stage ablation tests whether hyperparameter prediction benefits from deeper semantic features. We compare the default stage choice (0--1) against a single shallow stage (\emph{stage-0 only}) and the full backbone (\emph{stages 0--3}), keeping all other settings identical (same training subset, schedule, head, and seeds).

\begin{table}[!htbp]
\centering
\setlength{\tabcolsep}{6pt}
\caption{Backbone-stage ablation on the zero-shot mixed-noise setting (3 seeds). \emph{Stages $(0,1)$} is our default.}
\label{tab:ablation_stage}
\begin{tabular}{lcccc}
\toprule
Stage selection & \shortstack{Backbone\\params $\downarrow$} & PSNR $\uparrow$ & \shortstack{Gap to oracle\\(dB) $\downarrow$} & \shortstack{Param.\\rel.\ err.\ $\downarrow$} \\
\midrule
Stage~0 only & $0.42$M & $24.06 \pm 0.26$ & $0.46 \pm 0.26$ & $0.174$ \\
Stages $(0,1)$ (\textbf{ours}) & $2.18$M & $\mathbf{24.29 \pm 0.09}$ & $\mathbf{0.23 \pm 0.09}$ & $\mathbf{0.160}$ \\
\shortstack[l]{Full backbone\\(stages 0--3)} & $87.69$M & $23.96 \pm 0.06$ & $0.56 \pm 0.06$ & $0.316$ \\
\bottomrule
\end{tabular}
\end{table}

Stages $0$ and $1$ give the best result: removing stage~$1$ lowers PSNR by $0.23$\,dB, while adding the deeper semantic stages $2$ and $3$ lowers PSNR by $0.33$\,dB with more than 40 times as many backbone parameters. This pattern is consistent with hyperparameter prediction relying on low-level noise and edge statistics rather than semantic content. Relative parameter error also nearly doubles from $0.160$ to $0.316$ when deep stages are included, indicating that including deeper stages lowers performance under this training setup.

\clearpage
\section{Computational cost and classical comparisons}
\label{app:computational_cost}

\subsection{Hardware and oracle-labeling cost}

Predictor training and evaluation run on $1$--$5$ NVIDIA RTX 4090 GPUs (24\,GB VRAM each) in parallel across seeds and configurations, with one card per run. Source pretraining takes $\sim\!2.5$ hours per seed. DiffPIR few-shot finetuning takes $\le 1$ minute for $\le 32$ target labels and up to $\sim\!30$ minutes for $128$ target labels for each seed and method.

Oracle-label generation is the costly stage. TV/TGV source labels are generated on CPUs, with each image searched independently; in the paper run, the full $13$-source search was distributed across a shared CPU cluster with approximately $600$ CPU cores available and about 2\,GB RAM per core, over roughly $10$ days. DiffPIR target-label generation for the $4344$ WED $256{\times}256$ training images is GPU-bound; in the paper run, these labels were generated on a heterogeneous, dynamically scheduled pool of $8$--$16$ NVIDIA L40S and A100 GPUs over roughly $2$ days.

The classical-method comparison (Table~\ref{tab:classical}) times HyperDn and SURE on the same machine. HyperDn selects parameters with one GPU forward pass; the downstream TV solver runs on CPU. SURE and the oracle sweep both require many CPU TV solver calls per image.

After fixing $\zeta=1$ using the shared training-set calibration, generating $2$ image-specific $\lambda$ labels for predictor fitting requires $40$ DiffPIR reconstructions at $20$ candidates per image, versus $1{,}280$ for Afkham's $64$ labels. At roughly $10$ seconds per reconstruction, generating these labels would take $0.11$ and $3.56$ GPU-hours if run sequentially, respectively.

\subsection{Batched predictor throughput}

We time the GPU forward pass in single precision (FP32), processing each test set as one batch of equally sized images. The three resolutions are: WED $256{\times}256$ center crops, CBSD68 at $321{\times}481$, and Kodak24 at $512{\times}768$. For timing only, we rotate CBSD68 and Kodak24 portrait images $90^\circ$ to landscape so that all images in each dataset have the same shape. Table~\ref{tab:throughput_appendix} reports single-batch timing, throughput, and peak GPU memory for each set.

At $256{\times}256$, the forward pass takes $0.78$\,ms per image ($1280$\,img/s). At $512{\times}768$, throughput falls to $212$\,img/s, approximately in inverse proportion to pixel count across these measurements. Per-batch wall-clock time for the entire test set is $0.11$--$0.31$\,s in all three cases. These times measure only the GPU forward pass. Table~\ref{tab:classical} instead measures parameter-selection time at batch size $1$, including CPU$\to$GPU transfer and Python overhead.

\begin{table}[!htbp]
\centering
\setlength{\tabcolsep}{4pt}
\caption{HyperDn single-batch throughput (FP32; time and memory $\downarrow$, throughput $\uparrow$). Batch time is mean$\pm$population SD across $24$ forward passes after $3$ warmup passes, with CUDA synchronization. Each row uses one batch of size $bs$ and equally sized images. ``Mem/img'' divides peak forward memory by $bs$, so fixed model weights are shared across the batch.}
\label{tab:throughput_appendix}
\begin{tabular}{lccrrrr}
\toprule
Dataset & Shape & $bs$ & \shortstack{Batch time\\(ms)} & \shortstack{Per-image\\(ms)} & \shortstack{Throughput\\(img/s)} & Mem/img \\
\midrule
\shortstack[l]{WED\\(center-cropped)} & $256{\times}256$ & $400$ & $312.5\!\pm\!6.8$ & $0.78$ & $1280$ & $ 31$\,MB \\
CBSD68  & $321{\times}481$ &  $68$ & $124.4\!\pm\!0.8$ & $1.83$ & $ 546$ & $ 73$\,MB \\
Kodak24 & $512{\times}768$ &  $24$ & $113.1\!\pm\!1.0$ & $4.71$ & $ 212$ & $187$\,MB \\
\bottomrule
\end{tabular}
\end{table}

\clearpage

\subsection{Comparison with classical parameter selection}
\label{sec:exp_classical}

We compare HyperDn with Monte-Carlo Stein's unbiased risk estimate (SURE)~\citep{ramani2008sure} for per-image selection of $\lambda$ in L2-TV denoising under additive Gaussian noise.

\paragraph{Setup.}
All methods operate on the same three out-of-distribution test datasets (CBSD68 $321{\times}481$, Kodak24 $512{\times}768$, Urban100 $\sim\!1024{\times}768$) with identical per-image Gaussian noise levels $\sigma$:
\begin{itemize}
    \item \textbf{Avg-oracle (fixed $\bar{\lambda}$):} applies the arithmetic mean of the $20\,000$ STL10 L2-TV Gaussian training labels to every test image;
    \item \textbf{SURE}~\citep{ramani2008sure}: minimizes Monte-Carlo SURE over $15$ logarithmically spaced $\lambda$ values in $[10^{-5},10^{0.5}]$, using the true noise level $\sigma$. One standard-normal probe per image is fixed across candidates, with perturbation scale $10^{-3}$ and a seed derived from the observation's noise seed;
    \item \textbf{HyperDn:} a single forward pass of the source-pretrained HyperDn model;
    \item \textbf{Oracle:} per-image hierarchical grid search using the clean reference image.
\end{itemize}

\paragraph{Results.}
Table~\ref{tab:classical} reports PSNR, gap to oracle, and per-image wall-clock cost across three out-of-distribution test datasets spanning $321{\times}481$ (CBSD68) to $\sim\!1024{\times}768$ (Urban100). The non-adaptive avg-oracle baseline trails the per-image oracle by $1.19$--$1.58$\,dB on average, showing the limitations of this training-mean fixed-$\lambda$ baseline on these datasets.

HyperDn matches or outperforms SURE in quality on every dataset: its gap to oracle is lower on all three datasets (CBSD68: $0.01$ vs.\ $0.06$, Kodak24: $0.27$ vs.\ $0.32$, Urban100: $0.02$ vs.\ $0.08$\,dB). It is also much cheaper per image. SURE selects $\lambda$ by evaluating a risk estimate over many candidate values; each candidate requires a TV solve on the observation and another on its perturbed copy. Its measured per-image parameter-selection time is $2.1$\,s on CBSD68, $5.7$\,s on Kodak24, and $11.2$\,s on Urban100. HyperDn selects $\lambda$ with one forward pass ($14$\,ms at $321{\times}481$, $19$\,ms at $512{\times}768$, $47$\,ms at $\sim\!1024{\times}768$) and then invokes the same TV solver once. Thus the learned predictor replaces SURE's per-image search with one learned prediction while retaining near-oracle reconstruction quality.

\begin{table}[!htbp]
\centering
\setlength{\tabcolsep}{5pt}
\caption{Classical comparison on L2-TV Gaussian across three out-of-distribution test datasets. Per-image $\sigma$ is sampled from the STL10 training range. ``Mean gap to oracle'' is oracle PSNR minus method PSNR. ``Inference'' reports the per-image $\lambda$-selection cost. Each method then uses the same TV solver for final reconstruction, whose cost is excluded. HyperDn is reported at batch $1$ to match SURE's per-image $\lambda$ search.}
\label{tab:classical}
\begin{tabular}{llccc}
\toprule
Dataset & Method & Mean PSNR $\uparrow$ & \shortstack{Mean gap to oracle\\(dB)} $\downarrow$ & Inference $\downarrow$ \\
\midrule
\multirow{3}{*}{CBSD68 ($321{\times}481$)}
  & Avg-oracle                      & $25.73$ & $1.19$ & --- \\
  & SURE                            & $26.85$ & $0.06$ & $\sim\!2100$\,ms \\
  & \textbf{HyperDn}                & $\mathbf{26.91\!\pm\!0.0029}$ & $\boldsymbol{0.01\!\pm\!0.0029}$ & $\mathbf{14}$\,\textbf{ms} \\
\midrule
\multirow{3}{*}{Kodak24 ($512{\times}768$)}
  & Avg-oracle                      & $26.44$ & $1.58$ & --- \\
  & SURE                            & $27.70$ & $0.32$ & $\sim\!5720$\,ms \\
  & \textbf{HyperDn}                & $\mathbf{27.75\!\pm\!0.0030}$ & $\boldsymbol{0.27\!\pm\!0.0030}$ & $\mathbf{19}$\,\textbf{ms} \\
\midrule
\multirow{3}{*}{Urban100 ($\sim\!1024{\times}768$)}
  & Avg-oracle                      & $24.49$ & $1.37$ & --- \\
  & SURE                            & $25.78$ & $0.08$ & $\sim\!11220$\,ms \\
  & \textbf{HyperDn}                & $\mathbf{25.84\!\pm\!0.0045}$ & $\boldsymbol{0.02\!\pm\!0.0045}$ & $\mathbf{47}$\,\textbf{ms} \\
\bottomrule
\end{tabular}
\end{table}

\clearpage
\section{Predictor transfer: definitions and proof}
\label{app:transfer_theory}

\subsection{Outputs and data}
\label{sec:transfer_definitions}
We use the same fixed predictor $f$ to analyze target-task error between encoded hyperparameter
predictions and oracle labels. Each task fixes the denoising configuration, the rules for supplying task
information, and the oracle procedure, including search rules, solver settings, random seeds, and tie-breaking.
A sample $x$ includes the observed image, any per-image observation metadata, and reference information
for its oracle label; the predictor reads only the image and supplied task information. The distributions $P_1,\ldots,P_K,P_T$ share a
measurable sample space allowing different image sizes. For source $k$, $X_S^{(k)}\sim P_k$ and
$X_T\sim P_T$. Each source prediction and oracle procedure is defined on source and target samples;
the target procedure is defined on target samples. Thus, $q_k^\star(X_T)$ applies source task $k$'s
oracle procedure to the target sample and reference information.

For a task with $m\ge1$ active encoded coordinates, retain those coordinates in the padded slot order
of Section~\ref{sec:method}, set inactive coordinates to zero, and divide both the prediction and label
by $\sqrt m$. The vectors $f_k(x),f_T(x),q_k^\star(x),q_T^\star(x)$ denote these normalized outputs
in the common space $\mathbb R^{L_{\max}}$. Their squared Euclidean distance equals the mean squared
prediction error over the task's active encoded coordinates. Encoding, slot order, and normalization
are fixed throughout. Both discrepancies compare prediction errors in these encoded coordinates; their
interpretation follows the specified order and scales, including when tasks have different parameter meanings.

All prediction and oracle maps are measurable and finite-valued. Assume
$\varepsilon_S(f),\varepsilon_T(f)<\infty$ and
$\E_{P_T}\|f_k(X_T)-q_k^\star(X_T)\|^2<\infty$ for every source $k$.
For each $k$, $\Pi(P_k,P_T)$ is the set of joint distributions of $(X_S^{(k)},X_T)$ with marginals
$P_k,P_T$. For generalization across image distributions under the source tasks, the data term measures changes
in the predictor's errors, taking the infimum over this set. For generalization to the target task, the
task term measures error changes on the same target sample.
The weights $\alpha_k$ specify the pooled evaluation average in Eq.~\eqref{eq:predictor_transfer_risk}.

\subsection{Proof}
\label{sec:predictor_transfer_proof}
\begin{proof}[Proof of Theorem~\ref{thm:transfer_decomposition}]
Choose $\pi_k\in\Pi(P_k,P_T)$ for each source $k$. Choose source $k$ with probability $\alpha_k$, then draw
$(X_S^{(k)},X_T)$ from $\pi_k$.
For every selected source, adding and subtracting the intermediate predictions and labels gives
\begin{align*}
 f_T(X_T)-q_T^\star(X_T)
 ={}&f_k(X_S^{(k)})-q_k^\star(X_S^{(k)})\\
 &+\bigl([f_k(X_T)-f_k(X_S^{(k)})]-[q_k^\star(X_T)-q_k^\star(X_S^{(k)})]\bigr)\\
 &+\bigl([f_T(X_T)-f_k(X_T)]-[q_T^\star(X_T)-q_k^\star(X_T)]\bigr).
\end{align*}
All three terms have finite second moments. Take the $L^2$ norm, the square root of the expected
squared Euclidean norm, under this sampling law. Each pairing has target marginal $P_T$ and
$\sum_k\alpha_k=1$, so the left-hand side has norm $\varepsilon_T(f)$. The triangle inequality gives
\begin{align*}
 \varepsilon_T(f)\le{}&\varepsilon_S(f)+\Delta_{\rm task}(f)\\
 &+\left(\sum_k\alpha_k\E_{\pi_k}\bigl\|
 [f_k(X_T)-f_k(X_S^{(k)})]-[q_k^\star(X_T)-q_k^\star(X_S^{(k)})]\bigr\|^2\right)^{1/2}.
\end{align*}
Taking the infimum over $\pi_1,\ldots,\pi_K$ gives Eq.~\eqref{eq:transfer_decomposition_main}.
\end{proof}

\end{document}